\documentclass{article}

\PassOptionsToPackage{numbers, compress}{natbib}

    \usepackage[final]{neurips_2022}

\usepackage[utf8]{inputenc} 
\usepackage[T1]{fontenc}    
\usepackage{xr-hyper}
\usepackage{hyperref}       
\usepackage{url}            
\usepackage{booktabs}       
\usepackage{amsfonts}       
\usepackage{nicefrac}       
\usepackage{microtype}      
\usepackage{xcolor}         
\usepackage{booktabs} 
\usepackage{yfonts} 
\usepackage{algorithm}
\usepackage[noend]{algpseudocode}
\usepackage{microtype}
\usepackage{graphicx}
\usepackage{subfigure}
\usepackage{booktabs} 

\usepackage{amsmath}
\usepackage{amssymb}
\usepackage{mathtools}
\usepackage{amsthm}
\usepackage{hyperref}
\usepackage{lipsum}
\usepackage{caption}
\usepackage{xr}
\usepackage{float}
\usepackage{graphicx}
\usepackage{wrapfig}
\usepackage{soul}
\usepackage{gensymb}
\usepackage{bm}
\usepackage{upgreek}
\usepackage{appendix}
\usepackage{xcolor,colortbl} 
\usepackage{array,multirow}
\definecolor{omid_color}{RGB}{0 0 0}
\definecolor{removed_material}{RGB}{250 0 50}
\definecolor{unresolved}{RGB}{250 150 0}

\definecolor{navid_color}{RGB}{50 200 100}
\definecolor{revise_color}{RGB}{200 50 150}

\title{Autoinverse: Uncertainty Aware Inversion of Neural Networks}

\author{%
  Navid Ansari\\
  Max Planck Institute for Informatics\\
  Saarbrücken, Germany \\
  \texttt{nansari@mpi-inf.mpg.de} \\
  \And
    Hans-Peter Seidel\\
  Max Planck Institute for Informatics\\
  Saarbrücken, Germany \\
  \texttt{hpseidel@mpi-sb.mpg.de} \\
  \And
   Nima Vahidi Ferdowsi\\
  Max Planck Institute for Informatics\\
  Saarbrücken, Germany \\
  \texttt{nvahidi@mpi-inf.mpg.de} \\
  \And
  Vahid Babaei\\
  Max Planck Institute for Informatics\\
  Saarbrücken, Germany \\
  \texttt{vbabaei@mpi-inf.mpg.de} \\
}

\begin{document}

\maketitle

\begin{abstract}

\begin{itemize}
Neural networks are powerful surrogates for numerous forward processes. The inversion of such surrogates is extremely valuable in science  and engineering. The most important property of a successful neural inverse method is the performance of its solutions when deployed in the real world, i.e., on the \emph{native forward process} (and not only the learned surrogate). We propose \texttt{Autoinverse}, a highly automated approach for inverting neural network surrogates. Our main insight is to seek inverse solutions in the vicinity of reliable data which have been sampled form the forward process and used for training the surrogate model. \texttt{Autoinverse} finds such solutions by taking into account the predictive uncertainty of the surrogate and minimizing it during the inversion.
Apart from high accuracy, {\texttt{Autoinverse}} enforces the feasibility of solutions, comes with embedded regularization, and is initialization free. We verify our proposed method through addressing a set of real-world problems in control, fabrication, and design.
Our code and data are available at: https://gitlab.mpi-klsb.mpg.de/nansari/autoinverse
\end{itemize}

\end{abstract}
\section{Introduction} \label{sec:intro}
``... optimizing for the wrong thing — offering prayers to the bronze snake of data rather the larger force behind it.'' \cite{christian2016algorithms} 

With the deep learning breakthrough during the last decade, data-driven modeling using neural network based \textit{surrogates} is now a standard practice in science and engineering. 
These surrogates strive to imitate the behavior of a \emph{native forward process} (NFP), such as a physics simulation, which maps a \emph{design} into its \emph{performance}\footnote{{In the applications showcased in this paper (fabrication-oriented design and robotics), the term \textit{design} refers to the space where the input to the forward process is characterized and parameterized and \textit{performance} refers to the parameterized space of desired properties.} Commonly, \textit{hidden state} or \textit{parameters}, and \textit{measurement} or \textit{goal} are used interchangeably with design and performance, respectively}. 
While forward processes are essential for troubleshooting and analysis, oftentimes their ultimate application lies in their inversion, i.e., the reverse process of mapping performances into designs. 
%
%
Despite the recent progress in inversion of neural networks within multiple disciplines \cite{tung2017adversarial, jiang2021deep, gavriil2020computational, wu2015galileo}, a fundamental unaddressed question is still standing out. 
Inversion of a surrogate model, fitted to the data sampled from the NFP, is ultimately different than the inversion of the NFP itself. 
The source of this gap could be explicit, such as the noise in measurements, or implicit, such as the poor sampling of the NFP. 
Although the obtained solutions from inverting the surrogate can be evaluated on the NFP, none of the current neural inversion methods offers a tailored solution for addressing this important gap. 
%


%
Our main insight in this work is to \textit{expect} and \textit{account} for any potential mismatch between the data, and consequently the surrogate, on the one hand and the NFP on the other. %
Our proposed method, \texttt{Autoinverse}, realizes this vision by taking into account the predictive uncertainty of the surrogate and minimizing it during the inversion. 
Therefore, the inverted solutions avoid the {unreliable} regions within the training data. 
%
%

We show that our \texttt{Autoinverse} strategy can augment existing neural inversion methods (both optimization-based and architecture-based approaches) with uncertainty compensation in a simple and practical manner. 
\texttt{Autoinverse} closes the gap between the surrogate and the NFP \textit{not} through attempting a perfect fit of the surrogate to the NFP, an onerous task, but by finding inverse solutions in the vicinity of the reliable training data where the surrogate and the NFP are most similar. 
Neural inverse methods equipped with \texttt{Autoinverse} outperform their counterparts significantly on both standard data sampled from the NFP and imperfect data, e.g., those corrupted by noise. 
Apart from high accuracy, \texttt{Autoinverse} enforces the feasibility of solutions, comes with embedded regularization (freeing the inversion approaches from hand-crafted regularizations based on domain knowledge), and is initialization-free. 
It achieves all these properties in a highly automated manner and only with a light, intuitive tuning.

\section{Related work} \label{sec:relate_work}
\paragraph{Neural Network Inversion}
We can divide neural network inversion approaches into two main categories. First, \textit{inverse architectures}, where we compute a network architecture that takes a given performance and maps it into a (distribution of) design(s). 
Second, \textit{direct optimization}~\cite{sun2021amortized}, where we optimize for a design such that it produces the desired performance. 
Although the simplest inverse architecture can be attempted by training a neural network in the reverse direction, it fails because of the one-to-many nature of the mapping. %
\emph{Tandem} networks use an inverse architecture by employing a pre-trained forward network in order to compute a consistent loss. %
The tandem approach has been developed independently across different disciplines \cite{liu2018training,shi2018deep,sun2021amortized,zhu2017unpaired} dating back (at least) to~\citet{tominaga1996color}. 
Many inverse architectures try to model the conditional posterior, $p(x|y)$, using variational methods \cite{ma2019probabilistic,kiarashinejad2020deep} based on (conditional) variational auto-encoders \cite{kingma2013auto}. 
These networks condition the design on the target performance and yield a distribution of solutions from which multiple samples could be drawn. 
\citet{kruse2021benchmarking} show that the invertible neural networks (INNs), built upon normalizing flows \cite{dinh2016density}, give the highest accuracy in terms of both {surrogate error} and design posterior compared to a wide range of inverse architectures. 
%

When using {direct optimization} methods, gradient-based optimizers can be readily used as the neural surrogate is differentiable. 
\citet{ren2020benchmarking} use stochastic gradient descent via backpropagation with respect to the design variables, and present a highly accurate and practical method. 
They benchmark their method, dubbed as \textit{neural adjoint} (\textit{NA}), against a set of inverse architectures and obtain significantly more accurate solutions. 
\citet{sun2021amortized} showed a similar approach except using a quasi-Newton method for optimization. 
\citet{ansari2021mixed} push forward in this direction by demonstrating that, for piecewise linear neural networks, e.g., those with ReLU activation, the direct optimization can be formulated as a mixed-integer linear program (MILP) and thus obtain \textit{globally} {optimal} solutions. 
While optimization methods are very accurate, their main disadvantage is their performance as they can be orders of magnitude slower than inverse architectures. 
%

\paragraph{Neural Networks and Predictive Uncertainty}
While neural networks are ubiquitous in almost all branches of natural sciences, their weakness at quantifying predictive uncertainty impedes their use in crucial applications.
Using a Bayesian formalism~\cite{bernardo2009bayesian}, Bayesian neural networks (BNNs)~\cite{neal2012bayesian,denker1990transforming,MachineLearningI}, given the training data and a prior over network's parameters, compute the posterior distribution of the parameters. 
Having computed the posterior, the predictive uncertainty can be computed. 
The inference step in computing BNNs is known to be computationally hard~\cite{jospin2020hands}. 
This explains the popularity of simpler methods for estimating predictive uncertainty, such as Monte Carlo dropout~\cite{gal2016dropout} and Deep Ensembles~\cite{lakshminarayanan2016simple}. 
Deep Ensembles strikes a good balance between simplicity and practicality on the one hand and predictive performance on the other hand (Section~\ref{sec:deep_ensembles}).  
One of the main advantages of the Deep Ensembles is its capability to predict aleatoric and epistemic uncertainty separately. 
Aleatoric and epistemic uncertainty carry different information regarding the surrogate \cite{kendall2017uncertainties} and considering both of them improves the quality of the neural inversion (Section \ref{sec:epistemic_aleatoric_uncertainty}).
Neural networks capable of predictive uncertainty are increasingly adopted in many different applications, such as reinforcement learning and active learning~\cite{gal2017deep}.  
As we shall see, neural inversion is yet another domain that takes advantage of this trend.   


\section{Method} \label{sec:method}

\texttt{Autoinverse} is an easy-to-implement technique for augmenting neural {inverse} methods with uncertainty awareness. 
\texttt{Autoinverse} achieves this goal by, first, training a surrogate capable of predictive uncertainty \cite{mackay1995probable}. 
%
Second, relying on this trained surrogate and using a novel inversion cost function, \texttt{Autoinverse} finds accurate designs with minimal uncertainty. 
We apply \texttt{Autoinverse} on two inverse methods belonging to the two main neural inversion categories, i.e., optimization- and architecture-based in Sections~\ref{sec:NA_UANA} and \ref{sec:tandem_uatandem}, respectively. 
As we will see in Section~\ref{sec:deep_ensembles}, we rely on established methods~\cite{lakshminarayanan2016simple} to train networks equipped with predictive uncertainty.

\subsection{Uncertainty aware neural adjoint (\texttt{UANA})} \label{sec:NA_UANA}

\begin{figure}
    \centering     
    \subfigure[]{
        \label{fig:UANA_arch}
        \includegraphics[width=0.45\textwidth]{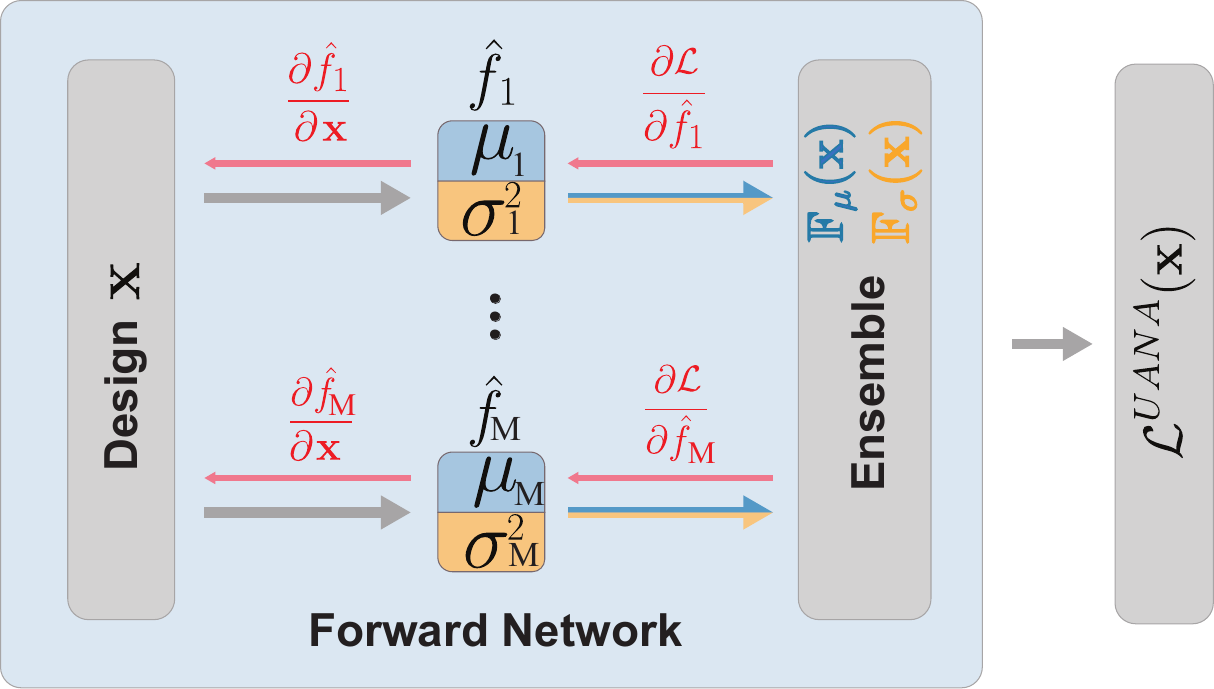}
        }
        \hfill
    \subfigure[]{
        \label{fig:UAT_arch}
        \includegraphics[width=0.50\textwidth]{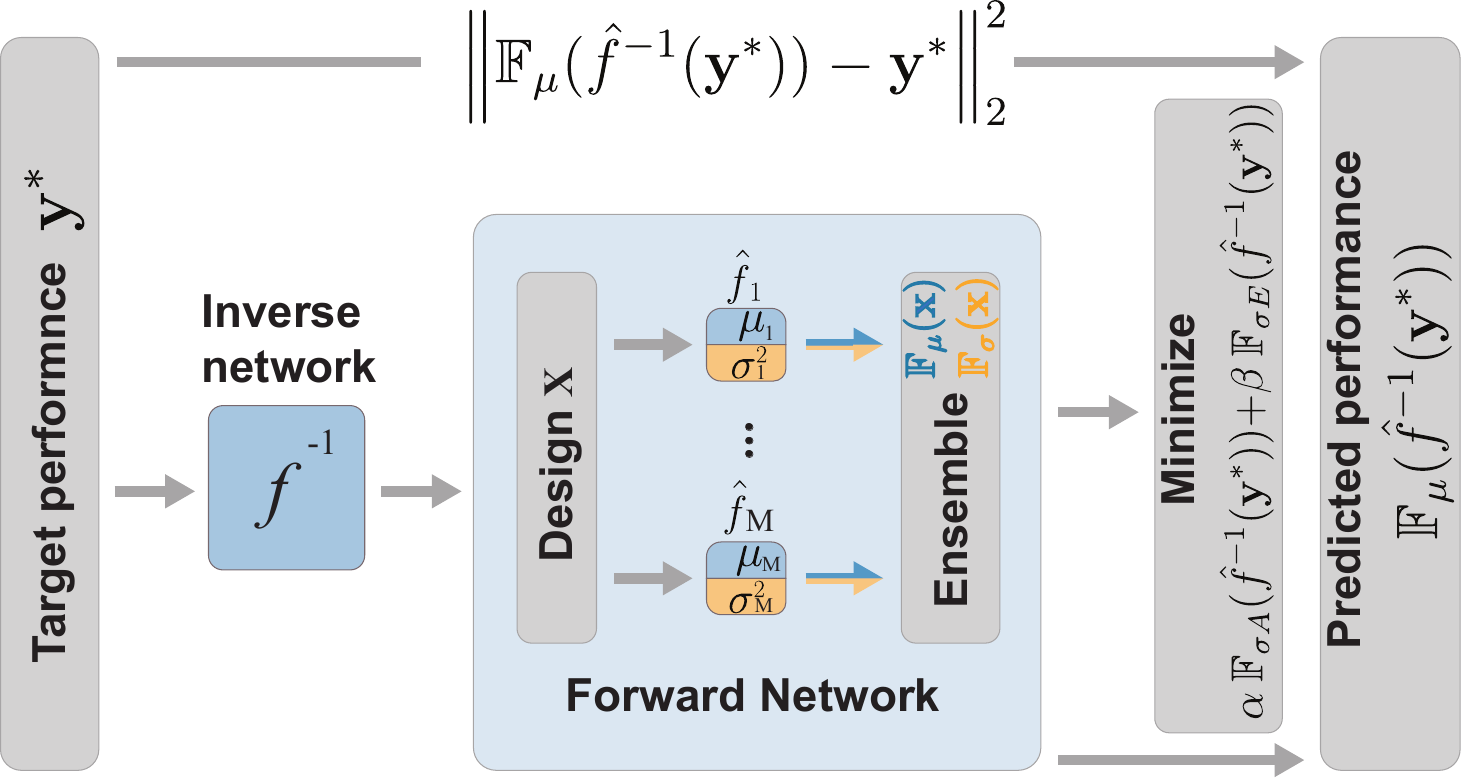}
        }
    \caption{By using a deep ensemble predictive uncertainty estimator as the forward model we can make conventional inversion methods uncertainty aware. On the left we can see \texttt{UANA} and on the right \texttt{UA-tandem} architecture.}
\end{figure}
Given a pretrained neural surrogate $\hat{f}(\cdot)$, neural adjoint (\texttt{NA})~\cite{ren2020benchmarking} is an inverse method that uses the cost function $\mathcal{L}^{NA}(\mathbf{\cdot})$ to push designs $\mathbf{x}$ to have a performance $\hat{f}(\mathbf{x})$  as close as possible to a desired performance $\mathbf{y^{*}}$: 

\begin{equation}
    \mathcal{L}^{NA}(\mathbf{x}) := \arg\min_{\mathbf{x}} \left\| \hat{f}(\mathbf{x}) - \mathbf{y^{*}} \right\|_{2}^{2}. 
\label{eq:NA_loss}
\end{equation}
\texttt{NA} uses gradient descent to iteratively reduce the cost function, a scheme much like training neural networks but with the input as the optimization variable instead of network's weights and biases. Equation~ \ref{eq:NA_grad} shows a single \texttt{NA} iteration with $\delta$ as the step size:
\begin{equation}
    \mathbf{x}^{z} = \mathbf{x}^{z-1} - \delta(\frac{\partial \mathcal{L}^{NA}}{\partial \hat{f}} \times \frac{\partial \hat{f}}{\partial \mathbf{x}}).
\label{eq:NA_grad}
\end{equation}

\texttt{Autoinverse} proposes to perform the inversion using a pretrained BNN. We use Deep Ensembles~\cite{lakshminarayanan2016simple} made of $M$ neural networks capable of a prediction $\mathbb{F}_{\mathbf{\mu}}(\mathbf{x})$, as well as its \textit{aleatoric} $\mathbb{F}_{\sigma A}(\mathbf{x})$ and \textit{epistemic} $\mathbb{F}_{\sigma E}(\mathbf{x})$ uncertainties (Section~\ref{sec:deep_ensembles}). 
Aleatoric uncertainty increases as the noise level in the training data increases.
Epistemic uncertainty measures the uncertainty in the model.
\texttt{Autoinverse} modifies \texttt{NA} such that we obtain solutions $\mathbf{x}$ that have performances close to the target performance $\mathbf{y^{*}}$ \textit{while} resulting in small aleatoric and epistemic uncertainties:
\begin{equation}
    \mathcal{L}^{UANA}(\mathbf{x}) := \ \arg\min_{\mathbf{x}} \left\| \mathbb{F}_{\mathbf{\mu}}(\mathbf{x}) -  \ \mathbf{y^{*}} \right\|_{2}^{2} + \alpha \ \mathbb{F}_{\sigma A}(\mathbf{x}) + \beta \ \mathbb{F}_{\sigma E}(\mathbf{x})
\label{eq:UANA_loss}
\end{equation}
We introduce $\alpha$ and $\beta$ as hyperparameters to adjust the relative significance of aleatoric and epistemic uncertainties, respectively.

Equation~\ref{eq:UANA_grad} shows how one iteration of \texttt{UANA} requires the back-propagation using the ensemble of all gradients of $M$ networks of Deep Ensembles:
\begin{equation}
    \mathbf{x}^{z} = \mathbf{x}^{z-1} - \delta\sum_{m=1}^{M}(\frac{\partial \mathcal{L}^{UANA}}{\partial \hat{f}_m} \times \frac{\partial \hat{f}_m}{\partial \mathbf{x}}) 
    \label{eq:UANA_grad}
\end{equation}
where $\hat{f}_m$ represents one of the networks in the ensemble.
Figure \ref{fig:UANA_arch} depicts this collective procedure where each individual network in the ensemble votes for the direction where updating the design will lead to the maximal accuracy and minimal uncertainty.

\subsection{Uncertainty aware tandem (\texttt{UA-tandem})} \label{sec:tandem_uatandem}

\texttt{Tandem} is a representative of architecture-based methods in which we train an inverse network $\hat{f}^{-1}(\cdot)$ in a manner resembling the encoder-decoder architecture (\cite{shi2018deep, sun2021amortized, tominaga1996color}).
Unlike the encoder-decoder approach, we start with training the forward model $\hat{f}(\cdot)$.
Then we freeze the trainable parameters of $\hat{f}(\cdot)$ and train the inverse model $\hat{f}^{-1}(\cdot)$ in the position of the encoder in order to decrease the cost function:
\begin{equation}
    \mathcal{L}^{T}(\hat{f}^{-1}(\mathbf{y^{*}})) := \arg\min_{\hat{f}^{-1}(\cdot)} \left\| \hat{f}(\hat{f}^{-1}(\mathbf{\mathbf{y^{*}}})) - \mathbf{y^{*}} \right\|_{2}^{2}.
\label{eq:tandem_loss}
\end{equation}
Once $\hat{f}^{-1}(\cdot)$ is trained we can simply query it to find designs with our desired performances: 
\begin{equation}
\hat{f}^{-1}(\mathbf{y^{*}}) = \mathbf{x}.
\label{eq:tandem_inverse}
\end{equation}
The uncertainty-aware tandem (\texttt{UA-tandem}) follows the same procedure except that it replaces $\hat{f}(\cdot)$ with $\mathbb{F}_{\mathbf{\mu}}(\cdot)$. Additionally, it includes the uncertainties in the loss:
\begin{equation}
    \mathcal{L}^{UAT}(\hat{f}^{-1}(\mathbf{y^{*}})) := \arg\min_{\hat{f}^{-1}(\cdot)} \left\| \mathbb{F}_{\mathbf{\mu}}(\hat{f}^{-1}(\mathbf{\mathbf{y^{*}}})) - \mathbf{y^{*}} \right\|_{2}^{2} +  \alpha \ \mathbb{F}_{\sigma A} (\hat{f}^{-1}(\mathbf{y^{*}})) + \beta \ \mathbb{F}_{\sigma E}(\hat{f}^{-1}(\mathbf{y^{*}})).
\label{eq:UAT_loss}
\end{equation}
Figure \ref{fig:UAT_arch} depicts the architecture of \texttt{UA-tandem}.

\subsection{Predictive uncertainty using Deep Ensembles} \label{sec:deep_ensembles}
\begin{wrapfigure}[6]{r}{0.20\textwidth}
  \vspace{-6mm}
  \centering
  \includegraphics[width=0.20\textwidth]{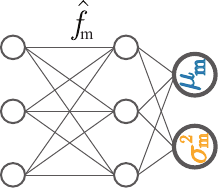}\\
  \label{fig:single_ensemble}
\end{wrapfigure}
Deep Ensembles comprises of an ensemble of $M$ neural networks $\hat{f}_{m}$ each capable of a prediction $\mu_{m}(\mathbf{x})$ and its associated uncertainty $\sigma_{m}(\mathbf{x})$ in form of a Gaussian distribution $\mathcal{N}(\mu_{m}(\mathbf{x}), \sigma_{m}(\mathbf{x}))$.
The cost function for training each network in the ensemble is the negative log likelihood~\cite{nix1994estimating}:
\begin{equation}
\mathcal{L}_{m}^{NLL} := \frac{\log (\sigma_{m}^{2}(\mathbf{x}))}{2} + \frac{(\mathbf{y^{*}} - \mu_{m}(\mathbf{x}))^{2}}{2\sigma_{m}^{2}(\mathbf{x})}.
\label{eq:NLL_loss}
\end{equation}
Intuitively, in case of the {aleatoric uncertainty} $\mu_{m}(\mathbf{x})$ fails to reliably predict $\mathbf{y}^{*}$.
Hence, $\sigma^{2}_{m}(\cdot)$ must increase to reduce the loss while the first term ensures $\sigma^{2}_{m}(\cdot)$ does not diverge to infinity.
The next step is to ensemble the results of all the networks into a single prediction and a single uncertainty.
Deep Ensembles~\cite{lakshminarayanan2016simple} models the ensemble as a single Gaussian distribution $\mathcal{N}(\mathbb{F}_{\mathbf{\mu}}(\mathbf{x}), \mathbb{F}_{\sigma}(\mathbf{x}) )$ approximating the mixture of $M$ previously computed Gaussian distributions

\begin{subequations} \label{eq:mixture}
\begin{gather}
\mathbb{F}_{\mathbf{\mu}}(\mathbf{x}):= \frac{1}{M} \ \sum_{m} \ \mu_{m}(\mathbf{x}) \label{eq:F_mu}, \\
\mathbb{F}_{\sigma}(\mathbf{x}) =  \frac{1}{M} \ \sum_{m} \ (\sigma^{2}_{m} (\mathbf{x}) + \ \mu_{m}^{2}(\mathbf{x}) )- \mathbb{F}_{\mathbf{\mu}}^{2}(\mathbf{x}). \label{eq:F_sigma}
\end{gather}
\end{subequations}

The uncertainty of the ensemble can be decomposed into two input-dependent uncertainties, i.e., aleatoric $\mathbb{F}_{\sigma A}(\mathbf{x})$ and {epistemic} $\mathbb{F}_{\sigma E}(\mathbf{x})$ through slight modification to Equation~\ref{eq:F_sigma}  \cite{kendall2017uncertainties}. 
\begin{subequations} \label{eq:sigma}
\begin{gather}
\mathbb{F}_{\mathbf{\sigma}}(\mathbf{x}) := \alpha \ \mathbb{F}_{\sigma A} (\mathbf{x}) + \beta \ \mathbb{F}_{\sigma E} (\mathbf{x}) \label{eq:sigma_all}, \\
\mathbb{F}_{\sigma A} (\mathbf{x}) := 
\frac{1}{M} \ \sum_{m} \ \sigma^{2}_{m}(\mathbf{x}) \label{eq:sigma_aleatoic},\\
\mathbb{F}_{\sigma E}(\mathbf{x}) := \frac{1}{M} \  \sum_{m} (\mu_{m}^{2}(\mathbf{x}) - \mathbb{F}_{\mathbf{\mu}}^{2}(\mathbf{x})) \label{eq:sigma_epistemic}.
\end{gather}
\end{subequations}

This enables us to control the behavior of the neural inversion by tuning the relative importance of these two uncertainties.
%
%

%

\section{Evaluation} \label{sec:evaluation}
We evaluate the performance of \texttt{Autoinverse} through experimenting with the existing neural inverse methods and their uncertainty-aware counterparts.  
In the paper, we focus on \texttt{NA} and \texttt{UANA} while \texttt{tandem} and \texttt{UA-tandem} are evaluated mainly in the Appendix. 
We perform the evaluation on a set of applications in robotics, and fabrication-oriented computational design. 
We report the error by running the experiments 3 times to capture the variations. 
More details are provided in each case study and in Appendix (Section \ref{sec:ste_tuning_imp_details}).

Equations~\ref{eq:UANA_loss} and~\ref{eq:UAT_loss} have two hyperparameters ($\alpha$ and $\beta$) that keep the balance between the MSE, the aleatoric uncertainty and the epistemic uncertainty.
We observe that with a relatively larger epistemic weight ($\beta$) we obtain better results.
Exploiting this intuition, we tune these parameters for 3 different sets of values for \{$\alpha$, $\beta$\}: $\left\{ \{0.1, 1\}, \{1, 10\}, \{10, 100\}\right\}$.
We then make two finer step depending on the outcome of the former evaluation and choose the best set of weights.
We keep this budget of 5 experiment runs for the rival methods as well. 
Typically, we use 10\% of the target performance for tuning our inverse methods.

\subsection{Experiments} \label{sec:experiments}
\paragraph{Multi-joint robot} is a simple inverse kinematics problem which is being used as a standard test for neural inverse problems \cite{kruse2021benchmarking, ren2020benchmarking}.
In this problem, the \emph{design} is the 1D position of the base of the multi-joint robot along with the 1D rotation of its three joints.
The inverse problem concerns finding a combination of positions and angles for the base and the joints such that the tip of the arm lands the target position, i.e., the \emph{performance}.
We follow \cite{ren2020benchmarking} for setting up this experiment and its corresponding analytical equation used as the NFP.
The training data consists of 10,000 pairs of samples generated by randomly sampling the NFP.
\paragraph{Spectral printer}
Spectral printing enables digital fabrication of the object's appearance faithfully (\cite{hersch2005improving, chen2004six}). 
Unlike reproduction of the color (e.g., RGB), reproducing the spectrum ensures that the original and the duplicate remain visually similar independent of the color of the light source. 
Spectral printing has various important applications specially in fine art reproduction using both 2D and 3D printing \cite{morovivc2012revisiting, ansari2020mixed, shi2018deep}. 
Deep neural networks are becoming the main computational tool for modeling the spectral printing process specially when dealing with a large number of inks. 
{The final objective is to find the correct ink densities at each pixel that can best estimate the target 31D spectrum.}

In this experiment, we create the NFP by simulating a printer using an ensemble of 20 neural networks. 
The design space comprises the ink densities and our \texttt{spectral printer} NFP predicts the spectrum of the resulting color.
We use real, measured data  from an 8-channel printer with 8 EPSON inks including Cyan (C), Magenta (M), Yellow (Y), Black (K), Light black (LK), Light light black (LLK), Light Cyan (LC), and Light Magenta (LM).
The light inks, added to the standard CMYK to improve the print quality, introduce significant multi-modality.
All networks in the ensemble NFP are trained on 40,000 printed patches \cite{ansari2020mixed} consisting of different ink-density combinations and their corresponding spectra. 
The ensembling is intended for an accurate NFP and is independent of our use of Deep Ensembles for computing the uncertainty. 
The visual nature of \texttt{spectral printer} makes it attractive for analyzing different methods.
We release this NFP to the public to add another example to the neural inversion testbed.

\paragraph{Soft robot}
are made of soft, flexible materials.
This unique property has made them suitable candidates for interaction with humans in applications like minimally invasive surgery and advanced prosthetics \cite{cianchetti2018biomedical}.
Unlike \texttt{multi-joint robot} with a limited number of rotating joints, each segment of the {soft robot} is a potential actuator that through their contraction and expansion can determine the robot's final shape.
This inverse kinematics problem is typically solved through partial differential equations (\cite{xue2020amortized}).
In order to accelerate the solve time of this inverse kinematics problem, \citet{sun2021amortized} proposed a neural surrogate modeling of the problem and its inversion via \texttt{tandem}.
The design space in this problem consists of the contraction or expansion of 40 controllable soft edge segments.
The superposition of all the actuations determines the final deformation position of the {soft robot} via the position of its 206 vertices.

We use an FEM-based simulation (\cite{xue2020amortized, hughes2012finite}) as our NFP.
We model the relationship between the actuations and the final shape of the {soft robot} with a neural-network surrogate.
Our goal is to solve the neural inversion to find a suitable set of actuations (design) that brings the tip of the {soft robot} to the target position (performance).
The training data consists of 50,000 samples queried by random sampling the actuation with an expansion ratios between -0.2 and 0.2 \cite{sun2021amortized}.
The designs are then evaluated on the FEM-based NFP to calculate their deformations. 

\subsection{Quantitative comparison of different neural inversion methods}
\label{sec:Quantitative-comparison}
%



	
\definecolor{lg}{gray}{0.92}
\definecolor{lgg}{gray}{0.975}
\newcolumntype{g}{>{\columncolor{lg}}c}
	\begin{table*}
		\centering
		\caption{The NFP and surrogate errors (mean $\pm$ STD) of different neural inverse methods on \texttt{multi-joint robot} for 1000 target locations.}
		\resizebox{\linewidth}{!}{%
			\begin{tabular}{cgggggggggggggg} 
				\toprule
				\cmidrule(lr){2-7}
				\rowcolor{white}
				Error & \texttt{NA} & \texttt{UANA} & \texttt{tandem} &\texttt{UA-tandem} & \texttt{MINI}& 
				\texttt{INN}\\
				\toprule


\rowcolor{lg}
 NFP& ($3.24 \pm 0.51$) & $\mathbf{(3.21 \pm 1.48)}$ & ($4.42 \pm 1.56$) & $\mathbf{(4.04 \pm 0.38) }$ & $1.6$ & ($9.48 \pm 0.021$)\\
 \rowcolor{lg}
 & $\times 10^{-4}$ & $\mathbf{\times 10^{-6}}$ & $\times 10^{-3}$ & $\mathbf{ \times 10^{-5}}$ & $\times 10^{-3}$  & $\times 10^{-3}$ \\
 \rowcolor{white}
Surrogate & $(1.99 \pm 0.05) $ & ($9.13 \pm 6.08$) & ($8.58 \pm 3.00$) & ($7.10 \pm 0.64$) & 0 &$(2.04 \pm 0.017)$\\
 \rowcolor{white}
 & $\times 10^{-8}$ & $\times 10^{-7}$ &$\times 10^{-6}$ & $\times 10^{-6}$ & &$ \times 10^{-13}$\\

				\bottomrule
			\end{tabular}
		}
		\label{tab:diverse_methods}
	\end{table*}
	
%
We evaluate the accuracy of a set of neural inversion methods on \texttt{multi-joint robot} in terms of both the NFP and the surrogate errors.
The surrogate error is the difference between the `re-prediction' of the obtained inverse solution by the surrogate neural network and the target performance.
The NFP error is the difference between the target and the performance of the generated design evaluated on the NFP.
In addition to the core methods described so far, we evaluate mixed-integer neural inversion (\texttt{MINI})~\cite{ansari2021mixed}, and the invertible neural network (\texttt{INN}) \cite{ardizzone2018analyzing}. 

Table~\ref{tab:diverse_methods} summarizes the inversion results on 1000 randomly sampled target locations for the multi-joint robotic arm. %
We keep our evaluation fair by setting the capacity of the neural surrogates comparable wherever possible. 
For instance, all methods except \texttt{MINI} have around $3$~million parameters (see Appendix, Table \ref{tab:networks-param-robotic-arm-config} for more details).
We also used equal computational resources for the tuning.
Methods with hyperparameters, like \texttt{UANA} and \texttt{UA-tandem}, are tuned in 5 stages by hand.
Alternatively, the methods without hyperparameters (\texttt{NA, tandem}) are given $5 \times$ extra budget for inversion:
We run \texttt{NA} and \texttt{tandem} $5 \times$ and choose the model that generates the best NFP error.
We repeat this process $3 \times$ and report the standard error.
\texttt{MILP} and \texttt{INN} are fundamentally different methods. 
\texttt{MILP} finds the global optimum and thus does not need tuning.
\texttt{INN} has a latent space which we can sample to generate diverse designs.
We sample INN's latent space 1024 times for all 1000 targets, evaluate them on the NFP, and report the best NFP error.
We repeat this process three times to generate the standard error.

Table~\ref{tab:diverse_methods} shows the outstanding accuracy of the inverse methods that adopt \texttt{Autoinverse}, i.e., \texttt{UANA} and \texttt{UA-tandem}, in terms of the NFP error. 
It also demonstrate how even a perfect surrogate error (e.g., \texttt{MINI}) does not guarantee accurate solutions when tested on the NFP.
A second look at Table~\ref{tab:diverse_methods} reveals further interesting insights. 
Although \texttt{NA} obtains notably lower surrogate error than its uncertainty-aware counterpart (\texttt{UANA}), it performs significantly worse in terms of the NFP error.
The main reason for this trend is that, when optimizing the surrogate, \texttt{UANA} is not only concerned with finding accurate designs leading to a small accuracy gap between the target and candidate performances, i.e., the surrogate error, but also with those designs featuring low uncertainty (through the uncertainty term in Equation~\ref{eq:sigma}). 
Therefore, \texttt{UANA} achieves high accuracy in terms of the essential NFP error at the cost of worsening the inconsequential surrogate error. 
Furthermore, \texttt{UA-tandem}, for example, achieves better performance than \texttt{NA}. 
This is a remarkable result for an architecture-based method given it is significantly faster than the optimization-based \texttt{NA}.
We report extensively the time performance and the details of the training for this and the next experiments in the Appendix. 
%
%

\subsection{Neural inversion in the presence of imperfect data}
\label{sec:epistemic_aleatoric_uncertainty}


\begin{table*}
		\centering
		\caption{The distribution of ink densities ($\geq 0.4$) after the inversion of \texttt{spectral printer} using \texttt{UANA}. Once we insert noise into LC channel or sample it sparsely, \texttt{Autoinverse} detects and avoids it. STD has been rounded to the nearest integer.}
		\resizebox{\linewidth}{!}{%
			\begin{tabular}{cgggggggggggggg} 
				\toprule

				\rowcolor{white}

				Model & dataset & NFP error &
				\cellcolor{cyan!70}C & \cellcolor{magenta!70}M & \cellcolor{yellow!70}Y & \cellcolor{black!50}K &
				\cellcolor{cyan!20} LC & \cellcolor{magenta!20} LM &
				\cellcolor{gray!30} LK & \cellcolor{gray!20} LLK\\
				\toprule


\rowcolor{white}
  &Standard& $(6.30 \pm 0.031) \times 10^{-3}$ & $186 \pm 2$ & $67 \pm 4$ & $63 \pm 6$ & $3 \pm 0$ &\cellcolor{green!30} $437 \pm 2$ & $356 \pm 7$ & $26 \pm 3$ & $348 \pm 5$\\

 \rowcolor{white}
 \texttt{\texttt{UANA}}&Sparse& $(5.64 \pm 0.017) \times 10^{-3}$ & $316 \pm 1$ & $60 \pm 2$ & $59 \pm 1$ & $2 \pm 0$ &\cellcolor{green!30} $\mathbf{0 \pm 0}$ & $326 \pm 5$ & $25 \pm 4$ & $323 \pm 11$\\

\rowcolor{white}
 &Noisy& $(6.13 \pm 0.026) \times 10^{-3}$ & $263 \pm 1$ & $67 \pm 4$ & $34 \pm 2$ & $1 \pm 0$ &\cellcolor{green!30} $\mathbf{0 \pm 0}$ & $276 \pm 3$ & $29 \pm 3$ & $378 \pm 5$\\

  \cmidrule(lr){2-11}
 
\rowcolor{white}
  &Standard& $(1.57 \pm 0.001) \times 10^{-1}$ & $1895 \pm 13$ & $1060 \pm 26$ & $1795 \pm 18$ & $162 \pm 11$ &\cellcolor{green!30} $865 \pm 8$ & $1396 \pm 30$ & $179 \pm 15$ & $2378 \pm 19$\\

 \rowcolor{white}
 \texttt{NA}&Sparse& $(1.34 \pm 0.007) \times 10^{-1}$ & $905 \pm 6$ & $606 \pm 7$ & $1515 \pm 8$ & $137 \pm 6$ &\cellcolor{green!30} $1604 \pm 11$ & $2130 \pm 17$ & $294 \pm 15$ & $2312 \pm 24$\\

\rowcolor{white}
 &Noisy& $(1.47 \pm 0.002) \times 10^{-1}$ & $1192 \pm 14$ & $988 \pm 10$ & $1128 \pm 10$ & $55 \pm 6$ &\cellcolor{green!30} $1029 \pm 23$ & $948 \pm 21$ & $283 \pm 12$ & $1742 \pm 37$\\
				\bottomrule
			\end{tabular}
		}

		\label{tab:ink_densities_distribution_(UA)NA}
\end{table*}

One of the main advantages of \texttt{Autoinverse} appears in scenarios where the training data suffers from noise, e.g., measurement noise or poor sampling, e.g., sparsity in some regions.
We evaluate the performance of \texttt{Autoinverse} on imperfect training data using \texttt{NA} and \texttt{UANA} (\texttt{tandem} and \texttt{UA-tandem} are evaluated under the same configuration in the Appendix Section \ref{sec:epistemic_aleatoric_uncertainty_sm}).
\begin{figure}[t] 
\centering
\includegraphics[width = 1\linewidth]{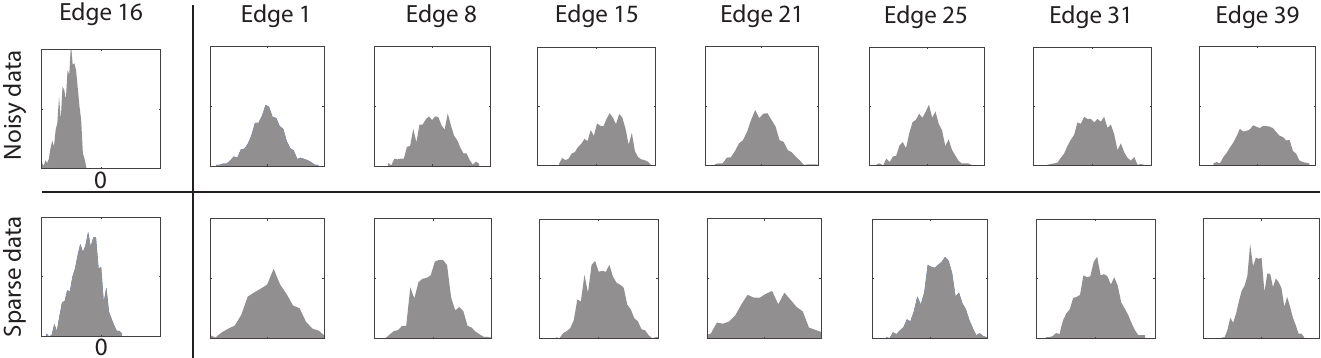}
\caption{Distribution of actuation values for \texttt{soft robot}. Edge 16 is corrupted by noise (top) and sampled sparsely (bottom) in positive range. While other randomly chosen edges feature both negative and positive actuations, \texttt{UANA} produces solutions without (or with less) positive actuations for Edge 16.}
\label{fig:soft_robot_aleatoric_epistemic}
\end{figure}
\paragraph{Locally sparse data}
We sample both \texttt{soft robot} and \texttt{spectral printer} NFPs (Section \ref{sec:experiments}) in a way that the data does not contain any samples from one of the inputs in a specific interval.
For \texttt{spectral printer}, we would like to find ink densities to reproduce the spectra of the colors in the painting in Figure~\ref{fig:valid_design} (made of 3568 distinct color spectra).
We sample the printer channels at 0 (no ink), 0.05, 0.1, 0.5, and 1 (full ink) densities to form the \textit{standard} training data (with no known uncertainty). 
We create a partially \textit{sparse} dataset similar to the standard one except for the Light Cyan (LC) channel for which we only have samples at 0, 0.05 and 0.1.
Table~\ref{tab:ink_densities_distribution_(UA)NA} shows that while for the standard dataset \texttt{UANA} finds inverse solutions that include the LC channel frequently (437 times), for the sparse dataset it avoids this channel  \textit{completely} and compensate for it using the Cyan channel.
\texttt{UANA} is able to avoid this channel as the epistemic uncertainty increases in sparse regions of dataset (see Appendix Section \ref{sec:SM_epistemic_aleatoric} for more details). 

In \texttt{soft robot} we sample the 16th (among 40) controllable edge only in the negative range (contraction only).
We then use the trained network to invert 1000 test samples.
Figure \ref{fig:soft_robot_aleatoric_epistemic} shows the distribution of each edge for 1000 inversion tasks.
We have plotted the distribution for the 16th edge as well as for 7 other randomly chosen edges.
As evident from Figure \ref{fig:soft_robot_aleatoric_epistemic} bottom row, \texttt{UANA} is highly reluctant to choose designs with positive actuations for this edge.
\paragraph{Locally noisy dataset} With the same problem configuration as before, we would like to test the robustness of \texttt{Autoinverse} on a dataset locally corrupted with noise.
We start with a standard dataset and inject Gaussian noise $\mathcal{N}(0,\, 0.1)\ $ to the spectrum of the samples with more than $0.4$ LC density.
Table \ref{tab:ink_densities_distribution_(UA)NA} shows how after introducing noise to the LC, the network avoids that channel and compensates it by using more Cyan instead.
In \texttt{soft robot} we corrupt the final shape of the soft robot with Gaussian noise $\mathcal{N}(0,\, 0.5)\ $ for all the shapes where the 16th edge has positive actuation.
As we can see in Figure~\ref{fig:soft_robot_aleatoric_epistemic} (top row), the inversion has completely avoided positive actuations for this edge.


\subsection{Autoinverse brings AutoML to neural inversion} \label{sec:valid_design}

\begin{figure}
    \centering

\includegraphics[width=0.5\textwidth]{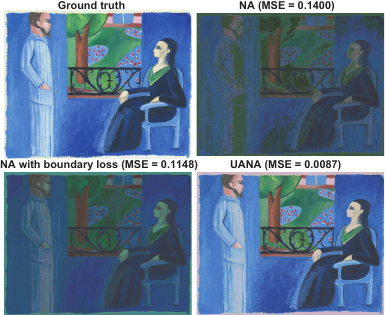}
\hfill
\includegraphics[width=0.3\textwidth]{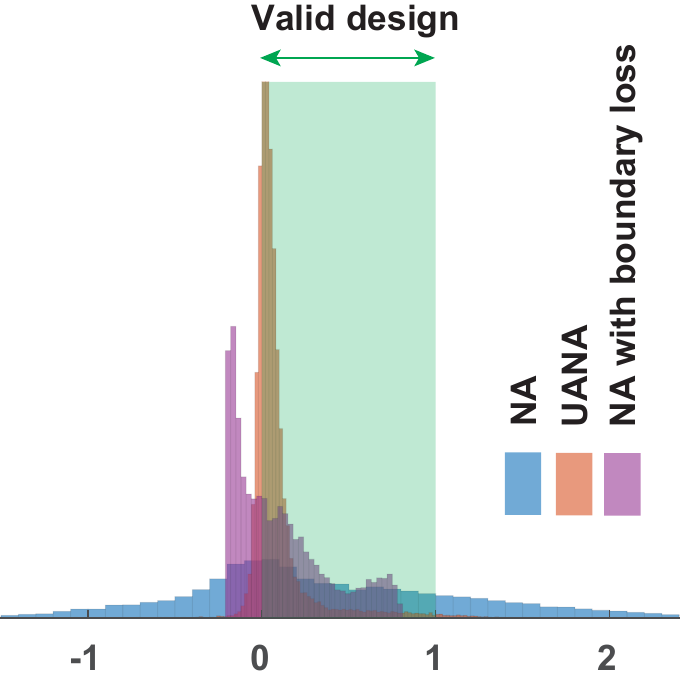}
\caption{In painting reproduction, \texttt{UANA} outperforms \texttt{NA} with and without a boundary loss term. The distribution of the inverse solutions is shown on right.}
  \label{fig:valid_design}
\end{figure}

\paragraph{Autoinverse incorporates feasibility.}
Deep Ensembles produces high epistemic uncertainty outside the distribution of the training data. 
This includes the regions where the NFP is not defined and thus not sampled. 
For example, ink densities outside $[0,1]$ are not printable. If such cases arise during the inversion, they are clipped to $[0,1]$~\cite{ansari2020mixed}.
In such regions, networks in the ensemble do not agree and thus the epistemic uncertainty increases. 
\texttt{Autoinverse} automatically avoids these unfeasible regions.
%
%
In order to simulate the result of the inversion, after clipping the out-of-range densities we feed them into the NFP.
As evident from Figure~\ref{fig:valid_design} left, inversion via NA results in a poor reproduction of the original painting.
%
%
\texttt{UANA}, on the other hand, achieves spectacular reproduction accuracy.  
This is explained by the plot in Figure \ref{fig:valid_design} right, showing the distribution of ink densities obtained from both methods.

\paragraph{Autoinverse has built-in regularization.}
We incorporate regularization into inversion methods in order to obtain solutions that, among other purposes, agree with the observations and follow a certain statistical distribution. 
Oftentimes, regularization is case-specific, requires human knowledge, and comes with unexpected side effects.
Here we show that \texttt{Autoinverse} follows the distribution of the training data naturally by taking into account the epistemic uncertainty.
We validate this point using both \texttt{spectral printer} and \texttt{soft robot} experiments. 
We show that \texttt{Autoinverse} without any explicit regularization performs better or on par with its counterpart inversion methods {equipped with} regularization. 

\begin{figure}[t] 
\centering
    \subfigure[]{
        \label{fig:soft_robot_shape}
        \includegraphics[width=0.48\textwidth]{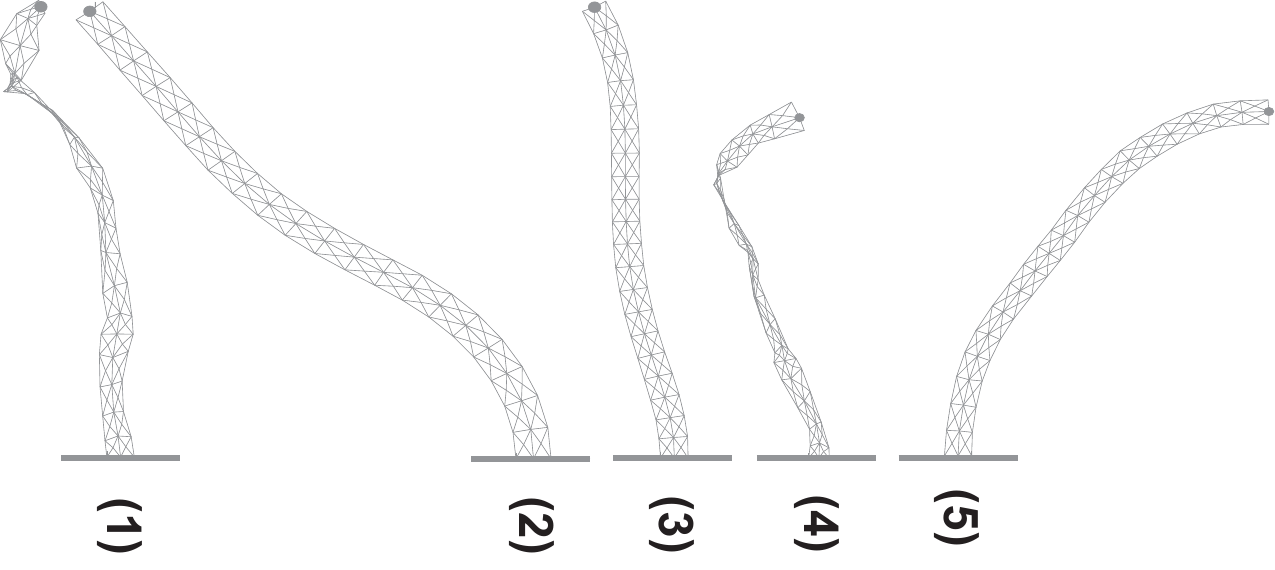}
        }
    \hfill
    \subfigure[]{
        \label{fig:sensetivity_init}
        \includegraphics[width=0.42\textwidth]{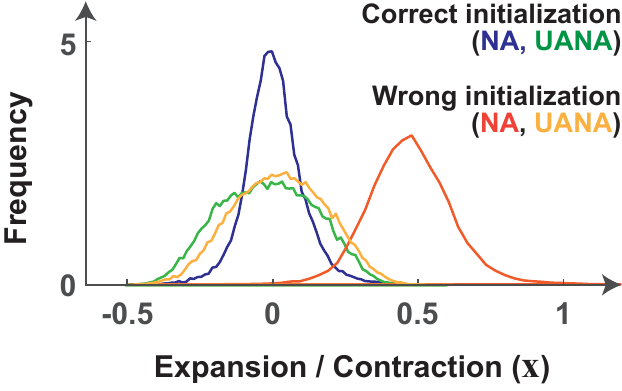}
        }

    \caption{On the left we can see the distribution of actuations calculated by \texttt{UANA} and \texttt{NA} with two different initialization: one initialization near the training data distribution and one far from it. On the right we can see a range of randomly chosen soft robot shapes calculated by different methods with different regularization and initialization.}%
    \label{fig:wrong_init}
\end{figure}

In the \texttt{spectral printer} experiment, we evaluate the effect of the \textit{boundary loss}, originally proposed as a generic regularization for \texttt{NA} to limit the designs within a box constraint (see \cite{ren2020benchmarking} and Appendix Section \ref{sec:SM_automl_neural_inversion}).
The boundary loss is added to Equation \ref{eq:NA_loss} and weighted using a hyperparameter. 
We tune this parameter with the same tuning budget we allocate for tuning the uncertainty weights (5 set of inversions on evaluation data).  
In Figure \ref{fig:valid_design}, we observe that although \texttt{NA} with boundary loss improves the distribution of ink densities within the valid region ($[0,1]$), it still trails the regularization-free \texttt{UANA} significantly.
Regularizing the \texttt{soft robot} is less intuitive as the superposition of all actuations determines the final shape and whether it is physically plausible.
In \cite{sun2021amortized}, the regularization is a smoothness term that keeps actuation values near each other (see Appendix Section \ref{sec:SM_automl_neural_inversion}).
Figure \ref{fig:soft_robot_shape}(1) demonstrates how \texttt{NA} fails without regularization to control the robot with a reasonable deformation.
Once the regularization is added to \texttt{NA}, the designs become physically meaningful (Figure \ref{fig:soft_robot_shape}(2)).
Figure \ref{fig:soft_robot_shape}(3) shows how \texttt{UANA} performs comparably without any regularization.
%

\paragraph{Autoinverse is initialization-free.}
Sensitivity to the initialization is a widely known issue in non-convex optimizations.
Despite equipping \texttt{NA} with the smoothness term in \texttt{soft robot}, Figure \ref{fig:sensetivity_init} demonstrates how an incorrect initialization can result in solutions with seemingly good surrogate and regularization loss (see Appendix Section \ref{sec:SM_automl_neural_inversion}) but in the infeasible region of the design space (Figure~\ref{fig:soft_robot_shape}(4)).
\texttt{UANA} with the same incorrect initialization \textit{and} without any regularization, leads to soft robot designs that reach the target location accurately and produce plausible deformations (Figure \ref{fig:soft_robot_shape}(5)). 
For \texttt{UANA}, the solver {starts with} reducing the main contributor to the optimization objective, i.e., the epistemic uncertainty.
Once a region with a reasonably small uncertainty is reached, the accuracy term starts to take effect and a desirable solution within the valid range of the design space is found. 

%

\subsection{Ablation studies}
\paragraph{Separating the ensembling effect}
Although in both \texttt{NA} and \texttt{UANA} we use surrogates with similar capacity (network size), one could argue that the higher performance of \texttt{UANA} comes from the ensemble architecture of its surrogate.
We detach the impact of ensembling and uncertainty awareness on the inversion performance by implementing \texttt{NA ensemble} where, instead of a single, large forward neural network, it uses an ensemble of networks.
The inversion procedure is identical to \texttt{NA} (see Appendix Section \ref{sec:Abelation_details} and Figure \ref{fig:NAEnsembleArchitecture}).
We employ \texttt{NA ensemble} in \texttt{multi-joint robot} with the same configurations as in Section~\ref{sec:Quantitative-comparison}.
The surrogate and NFP error for \texttt{NA ensemble} are $(3.30 \pm 0.59) \times 10^{-9}$ and $(1.17 \pm 0.32) \times 10^{-4}$, respectively.
Comparing these values with those in Table~\ref{tab:diverse_methods}, \texttt{NA ensemble} shows a slight improvement over \texttt{NA} but is significantly outperformed by \texttt{UANA}.
%

\paragraph{Diversity of activation functions in Deep Ensembles}
Deep Ensembles uses a similar set of networks with identical activation functions~\cite{lakshminarayanan2016simple}. 
In practice, we observe that a diverse set of activation layers leads to a better performance of \texttt{Autoinverse}. 
Different activation functions generate different behaviours, show higher disagreement where the training data is under-represented and, thus, result in {an accurate estimation of the landscape of} epistemic uncertainty~\cite{van2020uncertainty}. 
For all experiments in this work, we use a diverse range of activation functions (see Appendix Section \ref{sec:ste_tuning_imp_details}). 
As an ablation, we run \texttt{UANA} on \texttt{spectral printer} using sparse data (same configuration as Section~\ref{sec:epistemic_aleatoric_uncertainty}) but with ReLU as the only activation layer. 
In contrast to Table~\ref{tab:ink_densities_distribution_(UA)NA}, \texttt{UANA} (with ReLU only activation) does not completely avoid the sparse domain and delivers $19 \pm 4$ solutions that contain LC with densities $\geq 0.4$.


\section{Discussion} \label{sec:discussion}

The \texttt{Autoinverse} cost function is multi-objective. Instead of finding a single solution through a weighted combination of these objectives (what we have seen so far), we can capture the trade-off between the accuracy and uncertainty through computing the Pareto front (see Appendix Section \ref{sec:prato_front} for more details).
\texttt{Autoinverse} is an inversion \textit{strategy} that could be applied to various inverse methods, especially those that concern imitation of an original process (dubbed as NFP in our work). 
We look forward to see \texttt{Autoinverse} adopted for more inverse architectures beyond \texttt{tandem}, and more {direct optimizations} beyond first-order \texttt{NA}.

Predicting the uncertainty using Deep Ensembles~\cite{lakshminarayanan2016simple} is not the most efficient solution. This is because multiple forward networks are trained in two stages (once for the mean, and once jointly for the mean and variance). It is highly interesting to integrate more efficient uncertainty estimation methods, such as Monte Carlo dropout~\cite{gal2016dropout, kendall2017uncertainties}, into \texttt{Autoinverse}.

%
\begin{ack}
We are grateful to Xingyuan Sun and Szymon Rusinkiewicz who provided the repository of the soft robot simulation~\cite{sun2021amortized}, Sebastian Cucerca for the revision of the figures, and anonymous reviewers for valuable feedback. Special thanks to Jalil Maani and Soraya Iranpak for their kind support. 
\end{ack}
\bibliography{main}

\begin{thebibliography}{39}
\providecommand{\natexlab}[1]{#1}
\providecommand{\url}[1]{\texttt{#1}}
\expandafter\ifx\csname urlstyle\endcsname\relax
  \providecommand{\doi}[1]{doi: #1}\else
  \providecommand{\doi}{doi: \begingroup \urlstyle{rm}\Url}\fi

\bibitem[Ansari et~al.(2020)Ansari, Alizadeh-Mousavi, Seidel, and
  Babaei]{ansari2020mixed}
Navid Ansari, Omid Alizadeh-Mousavi, Hans-Peter Seidel, and Vahid Babaei.
\newblock Mixed integer ink selection for spectral reproduction.
\newblock \emph{ACM Transactions on Graphics (TOG)}, 39\penalty0 (6):\penalty0
  1--16, 2020.

\bibitem[Ansari et~al.(2021)Ansari, Seidel, and Babaei]{ansari2021mixed}
Navid Ansari, Hans-Peter Seidel, and Vahid Babaei.
\newblock Mixed integer neural inverse design.
\newblock \emph{arXiv preprint arXiv:2109.12888}, 2021.

\bibitem[Ardizzone et~al.(2019)Ardizzone, Kruse, Rother, and
  Köthe]{ardizzone2018analyzing}
Lynton Ardizzone, Jakob Kruse, Carsten Rother, and Ullrich Köthe.
\newblock Analyzing inverse problems with invertible neural networks.
\newblock In \emph{International Conference on Learning Representations}, 2019.
\newblock URL \url{https://openreview.net/forum?id=rJed6j0cKX}.

\bibitem[Bernardo and Smith(2009)]{bernardo2009bayesian}
Jos{\'e}~M Bernardo and Adrian~FM Smith.
\newblock \emph{Bayesian theory}, volume 405.
\newblock John Wiley \& Sons, 2009.

\bibitem[Chen et~al.(2004)Chen, Berns, and Taplin]{chen2004six}
Yongda Chen, Roy~S Berns, and Lawrence~A Taplin.
\newblock Six color printer characterization using an optimized cellular
  yule-nielsen spectral neugebauer model.
\newblock \emph{Journal of Imaging Science and Technology}, 48\penalty0
  (6):\penalty0 519--528, 2004.

\bibitem[Christian and Griffiths(2016)]{christian2016algorithms}
Brian Christian and Tom Griffiths.
\newblock \emph{Algorithms to live by: The computer science of human
  decisions}.
\newblock Macmillan, 2016.

\bibitem[Cianchetti et~al.(2018)Cianchetti, Laschi, Menciassi, and
  Dario]{cianchetti2018biomedical}
Matteo Cianchetti, Cecilia Laschi, Arianna Menciassi, and Paolo Dario.
\newblock Biomedical applications of soft robotics.
\newblock \emph{Nature Reviews Materials}, 3\penalty0 (6):\penalty0 143--153,
  2018.

\bibitem[Deb et~al.(2002)Deb, Pratap, Agarwal, and Meyarivan]{deb2002fast}
Kalyanmoy Deb, Amrit Pratap, Sameer Agarwal, and TAMT Meyarivan.
\newblock A fast and elitist multiobjective genetic algorithm: Nsga-ii.
\newblock \emph{IEEE transactions on evolutionary computation}, 6\penalty0
  (2):\penalty0 182--197, 2002.

\bibitem[Denker and LeCun(1990)]{denker1990transforming}
John~S Denker and Yann LeCun.
\newblock Transforming neural-net output levels to probability distributions.
\newblock In \emph{Proceedings of the 3rd International Conference on Neural
  Information Processing Systems}, pages 853--859, 1990.

\bibitem[Dinh et~al.(2016)Dinh, Sohl-Dickstein, and Bengio]{dinh2016density}
Laurent Dinh, Jascha Sohl-Dickstein, and Samy Bengio.
\newblock Density estimation using real nvp.
\newblock \emph{arXiv preprint arXiv:1605.08803}, 2016.

\bibitem[Gal and Ghahramani(2016)]{gal2016dropout}
Yarin Gal and Zoubin Ghahramani.
\newblock Dropout as a bayesian approximation: Representing model uncertainty
  in deep learning.
\newblock In \emph{international conference on machine learning}, pages
  1050--1059. PMLR, 2016.

\bibitem[Gal et~al.(2017)Gal, Islam, and Ghahramani]{gal2017deep}
Yarin Gal, Riashat Islam, and Zoubin Ghahramani.
\newblock Deep bayesian active learning with image data.
\newblock In \emph{International Conference on Machine Learning}, pages
  1183--1192. PMLR, 2017.

\bibitem[Gavriil et~al.(2020)Gavriil, Guseinov, P{\'e}rez, Pellis, Henderson,
  Rist, Pottmann, and Bickel]{gavriil2020computational}
Konstantinos Gavriil, Ruslan Guseinov, Jes{\'u}s P{\'e}rez, Davide Pellis, Paul
  Henderson, Florian Rist, Helmut Pottmann, and Bernd Bickel.
\newblock Computational design of cold bent glass fa{\c{c}}ades.
\newblock \emph{ACM Transactions on Graphics (TOG)}, 39\penalty0 (6):\penalty0
  1--16, 2020.

\bibitem[Hersch and Cr{\'e}t{\'e}(2005)]{hersch2005improving}
Roger~David Hersch and Fr{\'e}d{\'e}rique Cr{\'e}t{\'e}.
\newblock Improving the yule-nielsen modified neugebauer model by dot surface
  coverages depending on the ink superposition conditions.
\newblock In \emph{Color Imaging X: Processing, Hardcopy, and Applications},
  volume 5667, pages 434--447. SPIE, 2005.

\bibitem[Hughes(2012)]{hughes2012finite}
Thomas~JR Hughes.
\newblock \emph{The finite element method: linear static and dynamic finite
  element analysis}.
\newblock Courier Corporation, 2012.

\bibitem[Jiang et~al.(2021)Jiang, Chen, and Fan]{jiang2021deep}
Jiaqi Jiang, Mingkun Chen, and Jonathan~A Fan.
\newblock Deep neural networks for the evaluation and design of photonic
  devices.
\newblock \emph{Nature Reviews Materials}, 6\penalty0 (8):\penalty0 679--700,
  2021.

\bibitem[Jospin et~al.(2020)Jospin, Buntine, Boussaid, Laga, and
  Bennamoun]{jospin2020hands}
Laurent~Valentin Jospin, Wray Buntine, Farid Boussaid, Hamid Laga, and Mohammed
  Bennamoun.
\newblock Hands-on bayesian neural networks--a tutorial for deep learning
  users.
\newblock \emph{arXiv preprint arXiv:2007.06823}, 2020.

\bibitem[Kendall and Gal(2017)]{kendall2017uncertainties}
Alex Kendall and Yarin Gal.
\newblock What uncertainties do we need in bayesian deep learning for computer
  vision?
\newblock \emph{Advances in neural information processing systems}, 30, 2017.

\bibitem[Kiarashinejad et~al.(2020)Kiarashinejad, Abdollahramezani, and
  Adibi]{kiarashinejad2020deep}
Yashar Kiarashinejad, Sajjad Abdollahramezani, and Ali Adibi.
\newblock Deep learning approach based on dimensionality reduction for
  designing electromagnetic nanostructures.
\newblock \emph{npj Computational Materials}, 6\penalty0 (1):\penalty0 1--12,
  2020.

\bibitem[Kingma and Welling(2013)]{kingma2013auto}
Diederik~P Kingma and Max Welling.
\newblock Auto-encoding variational bayes.
\newblock \emph{arXiv preprint arXiv:1312.6114}, 2013.

\bibitem[Kruse et~al.(2021)Kruse, Ardizzone, Rother, and
  K{\"o}the]{kruse2021benchmarking}
Jakob Kruse, Lynton Ardizzone, Carsten Rother, and Ullrich K{\"o}the.
\newblock Benchmarking invertible architectures on inverse problems.
\newblock \emph{arXiv preprint arXiv:2101.10763}, 2021.

\bibitem[Lakshminarayanan et~al.(2016)Lakshminarayanan, Pritzel, and
  Blundell]{lakshminarayanan2016simple}
Balaji Lakshminarayanan, Alexander Pritzel, and Charles Blundell.
\newblock Simple and scalable predictive uncertainty estimation using deep
  ensembles.
\newblock \emph{arXiv preprint arXiv:1612.01474}, 2016.

\bibitem[Liu et~al.(2018)Liu, Tan, Khoram, and Yu]{liu2018training}
Dianjing Liu, Yixuan Tan, Erfan Khoram, and Zongfu Yu.
\newblock Training deep neural networks for the inverse design of nanophotonic
  structures.
\newblock \emph{ACS Photonics}, 5\penalty0 (4):\penalty0 1365--1369, 2018.

\bibitem[Ma et~al.(2019)Ma, Cheng, Xu, Wen, and Liu]{ma2019probabilistic}
Wei Ma, Feng Cheng, Yihao Xu, Qinlong Wen, and Yongmin Liu.
\newblock Probabilistic representation and inverse design of metamaterials
  based on a deep generative model with semi-supervised learning strategy.
\newblock \emph{Advanced Materials}, 31\penalty0 (35):\penalty0 1901111, 2019.

\bibitem[MacKay(1995)]{mackay1995probable}
David~JC MacKay.
\newblock Probable networks and plausible predictions-a review of practical
  bayesian methods for supervised neural networks.
\newblock \emph{Network: computation in neural systems}, 6\penalty0
  (3):\penalty0 469, 1995.

\bibitem[Michalski et~al.(1983)Michalski, Carbonell, and
  Mitchell]{MachineLearningI}
R.~S. Michalski, J.~G. Carbonell, and T.~M. Mitchell, editors.
\newblock \emph{Machine Learning: An Artificial Intelligence Approach, Vol. I}.
\newblock Tioga, Palo Alto, CA, 1983.

\bibitem[Morovi{\v{c}} et~al.(2012)Morovi{\v{c}}, Morovi{\v{c}}, Arnabat, and
  Garc{\'\i}a-Reyero]{morovivc2012revisiting}
Peter Morovi{\v{c}}, J{\'a}n Morovi{\v{c}}, Jordi Arnabat, and Juan~Manuel
  Garc{\'\i}a-Reyero.
\newblock Revisiting spectral printing: A data driven approach.
\newblock In \emph{Color and Imaging Conference}, volume 2012, pages 335--340.
  Society for Imaging Science and Technology, 2012.

\bibitem[Neal(2012)]{neal2012bayesian}
Radford~M Neal.
\newblock \emph{Bayesian learning for neural networks}, volume 118.
\newblock Springer Science \& Business Media, 2012.

\bibitem[Nix and Weigend(1994)]{nix1994estimating}
David~A Nix and Andreas~S Weigend.
\newblock Estimating the mean and variance of the target probability
  distribution.
\newblock In \emph{Proceedings of 1994 ieee international conference on neural
  networks (ICNN'94)}, volume~1, pages 55--60. IEEE, 1994.

\bibitem[Ren et~al.(2020)Ren, Padilla, and Malof]{ren2020benchmarking}
Simiao Ren, Willie Padilla, and Jordan Malof.
\newblock Benchmarking deep inverse models over time, and the neural-adjoint
  method.
\newblock \emph{arXiv preprint arXiv:2009.12919}, 2020.

\bibitem[Seitzer et~al.(2022)Seitzer, Tavakoli, Antic, and
  Martius]{seitzer2022pitfalls}
Maximilian Seitzer, Arash Tavakoli, Dimitrije Antic, and Georg Martius.
\newblock On the pitfalls of heteroscedastic uncertainty estimation with
  probabilistic neural networks.
\newblock \emph{arXiv preprint arXiv:2203.09168}, 2022.

\bibitem[Shi et~al.(2018)Shi, Babaei, Kim, Foshey, Hu, Sitthi-Amorn,
  Rusinkiewicz, and Matusik]{shi2018deep}
Liang Shi, Vahid Babaei, Changil Kim, Michael Foshey, Yuanming Hu, Pitchaya
  Sitthi-Amorn, Szymon Rusinkiewicz, and Wojciech Matusik.
\newblock Deep multispectral painting reproduction via multi-layer, custom-ink
  printing.
\newblock \emph{ACM Trans. Graph.}, 37\penalty0 (6):\penalty0 271:1--271:15,
  December 2018.

\bibitem[Sun et~al.(2021)Sun, Xue, Rusinkiewicz, and Adams]{sun2021amortized}
Xingyuan Sun, Tianju Xue, Szymon Rusinkiewicz, and Ryan~P Adams.
\newblock Amortized synthesis of constrained configurations using a
  differentiable surrogate.
\newblock In A.~Beygelzimer, Y.~Dauphin, P.~Liang, and J.~Wortman Vaughan,
  editors, \emph{Advances in Neural Information Processing Systems}, 2021.
\newblock URL \url{https://openreview.net/forum?id=wdIDt--oLmV}.

\bibitem[Tominaga(1996)]{tominaga1996color}
Shoji Tominaga.
\newblock Color control using neural networks and its application.
\newblock In \emph{Color Imaging: Device-Independent Color, Color Hard Copy,
  and Graphic Arts}, volume 2658, pages 253--260. International Society for
  Optics and Photonics, 1996.

\bibitem[Tung et~al.(2017)Tung, Harley, Seto, and
  Fragkiadaki]{tung2017adversarial}
Hsiao-Yu~Fish Tung, Adam~W Harley, William Seto, and Katerina Fragkiadaki.
\newblock Adversarial inverse graphics networks: Learning 2d-to-3d lifting and
  image-to-image translation from unpaired supervision.
\newblock In \emph{2017 IEEE International Conference on Computer Vision
  (ICCV)}, pages 4364--4372. IEEE, 2017.

\bibitem[Van~Amersfoort et~al.(2020)Van~Amersfoort, Smith, Teh, and
  Gal]{van2020uncertainty}
Joost Van~Amersfoort, Lewis Smith, Yee~Whye Teh, and Yarin Gal.
\newblock Uncertainty estimation using a single deep deterministic neural
  network.
\newblock In \emph{International conference on machine learning}, pages
  9690--9700. PMLR, 2020.

\bibitem[Wu et~al.(2015)Wu, Yildirim, Lim, Freeman, and
  Tenenbaum]{wu2015galileo}
Jiajun Wu, Ilker Yildirim, Joseph~J Lim, Bill Freeman, and Josh Tenenbaum.
\newblock Galileo: Perceiving physical object properties by integrating a
  physics engine with deep learning.
\newblock \emph{Advances in neural information processing systems}, 28, 2015.

\bibitem[Xue et~al.(2020)Xue, Beatson, Adriaenssens, and
  Adams]{xue2020amortized}
Tianju Xue, Alex Beatson, Sigrid Adriaenssens, and Ryan Adams.
\newblock Amortized finite element analysis for fast pde-constrained
  optimization.
\newblock In \emph{International Conference on Machine Learning}, pages
  10638--10647. PMLR, 2020.

\bibitem[Zhu et~al.(2017)Zhu, Park, Isola, and Efros]{zhu2017unpaired}
Jun-Yan Zhu, Taesung Park, Phillip Isola, and Alexei~A Efros.
\newblock Unpaired image-to-image translation using cycle-consistent
  adversarial networks.
\newblock In \emph{Proceedings of the IEEE international conference on computer
  vision}, pages 2223--2232, 2017.

\end{thebibliography}
\bibliographystyle{plainnat}

\newpage
{\LARGE\bfseries Appendix}
\appendix
\maketitle
\newcommand{\MultiJointRobotTrainingLoss}{%
\definecolor{lg}{gray}{0.92}
\definecolor{lgg}{gray}{0.975}
\newcolumntype{g}{>{\columncolor{lg}}c}
	\begin{table}

		\centering
		\caption{[Continued] Training details of different neural surrogate models used in inverse methods for \texttt{multi-joint robot}.}
                \small
			\begin{tabular}{cgggggggggggggg} 
				\toprule
				\rowcolor{white}
				Network's name & Sub-networks name & Total training time (s) & Total inversion time (s)& Training loss\\
				\toprule


\rowcolor{white}
 \texttt{INN}& - & - & $1024 \times (1.3 \times 10^{-2})$ & $2.10 \times 10^{-2}$ \\
 \rowcolor{lg}
 \texttt{NA}&- & 190 & 374& $3.33 \times 10^{-5}$ \\
\rowcolor{white}
 \texttt{NA ensemble} & single  & 1650 & 426 & $3.91 \times 10^{-6}$\\
 \rowcolor{lg}
 \texttt{UANA}&  $\mu$ networks & 1650 & 1075 & $2.34 \times 10^{-6}$\\
  \rowcolor{lg}
 &  $\sigma$ networks  & 192 & &\\
\rowcolor{white}
 \texttt{Tandem} & Forward network & 190 &  $3.8 \times 10^{-3}$ & $3.33 \times 10^{-5}$ \\
 \rowcolor{white}
&Inverse network & 181 & &\\
\rowcolor{lg} 
& $\mu$ networks& 1650  & & \\
 \rowcolor{lg}
 \texttt{UA-tandem}& $\sigma$ networks & 192 & $3.8 \times 10^{-3}$ & $2.34 \times 10^{-6}$ \\
 \rowcolor{lg}
&Inverse network  & 1109 & &\\
 \rowcolor{white}
 \texttt{MINI}&-& 150 & $2.98 \times 10^{4}$ & 4.89 $\times 10^{-4}$\\

				\bottomrule
			\end{tabular}
		
		\label{tab:networks-param-robotic-arm-loss}
	\end{table}
}

\newcommand{\MultiJointRobotConfig}{%
\definecolor{lg}{gray}{0.92}
\definecolor{lgg}{gray}{0.975}
\newcolumntype{g}{>{\columncolor{lg}}c}
	\begin{table*}
		\centering

		\caption{Training details of different neural surrogate models used in inverse methods for \texttt{multi-joint robot}.}
                \small
			\begin{tabular}{cgggggggggggggg} 
				\toprule
				\rowcolor{white}
				Network's name & Sub-networks name & Trainable parameters & Layer configuration\\
				\toprule


\rowcolor{white}
 \texttt{INN}& -  & 3727416 & \cite{ren2020benchmarking, ardizzone2018analyzing} \\
 \rowcolor{lg}
 \texttt{NA}&-& 3204302 & 100, 1000, 1500, 1000, 100 \\
\rowcolor{white}
 \texttt{NA ensemble} & Forward networks & 351802 $\times$ 10 & 100, 500, 500, 100\\
 \rowcolor{lg}
 \texttt{UANA}&  $\mu$ networks& 351802 $\times$ 10 & 100, 500, 500, 100\\
  \rowcolor{lg}
 &  $\sigma$ networks & 20902 $\times$ 10 & 100, 100, 100\\
\rowcolor{white}
 \texttt{Tandem} & Forward network & 3204302 &  100, 1000, 1500, 1000, 100 \\
 \rowcolor{white}
&Inverse network& 113804 & 100, 250, 250, 100\\
\rowcolor{lg}
& $\mu$ networks& 351802 $\times$ 10 & 100, 500, 500, 100\\
 \rowcolor{lg}
 \texttt{UA-tandem}& $\sigma$ networks & 20902 $\times$ 10 & 100, 100, 100 \\
 \rowcolor{lg}
&Inverse network& 117108 & 100, 250, 250, 100 \\
 \rowcolor{white}
 \texttt{MINI}&-& 10802 & 100, 100  \\

				\bottomrule
			\end{tabular}

		\label{tab:networks-param-robotic-arm-config}
	\end{table*}
}

\newcommand{\SpectralPrintingLoss}{%
\definecolor{lg}{gray}{0.92}
\definecolor{lgg}{gray}{0.975}
\newcolumntype{g}{>{\columncolor{lg}}c}
	\begin{table}
		\centering
		\caption{[Continued] Training details of different neural surrogate models used in inverse methods for \texttt{spectral printer}.}

                \small
			\begin{tabular}{cgggggggggggggg} 
				\toprule
				\rowcolor{white}
				Network's name & Sub-networks name & Training time & Inversion time & Training loss\\
				\toprule


 \rowcolor{lg}
 \texttt{NA}& -&  295 & 300 & $4.45 \times 10^{-6}$\\
\rowcolor{white}
 \texttt{NA ensemble} & Forward networks  & 441 & 563 & $3.38 \times 10^{-6}$\\
 \rowcolor{lg}
 \texttt{UANA}&  $\mu$ networks & 441 & $1.15\times 10^{3}$& $3.44 \times 10^{-6}$\\
  \rowcolor{lg}
 &  $\sigma$ networks & 240& &\\
\rowcolor{white}
 \texttt{Tandem}&Forward network & 295 & $1.07\times 10^{-2}$&$4.45 \times 10^{-6}$\\
 \rowcolor{white}
&Inverse network& 260 &  & &\\
\rowcolor{lg}
& $\mu$ networks& 441 & &\\
 \rowcolor{lg}
 \texttt{UATandem}& $\sigma$ networks & 240 & $1.04\times 10^{-2}$ & $3.44 \times 10^{-6}$\\
 \rowcolor{lg}
&Inverse network &  $1.08\times 10^{3}$ & &\\

				\bottomrule
			\end{tabular}

		\label{tab:networks_param_specral_printing-loss}
	\end{table}
	}

\newcommand{\SpectralPrintingConfig}{%
\definecolor{lg}{gray}{0.92}
\definecolor{lgg}{gray}{0.975}
\newcolumntype{g}{>{\columncolor{lg}}c}
	\begin{table}

		\centering
		\caption{Training details of different neural surrogate models used in inverse methods for \texttt{spectral printer}.}

                \small
			\begin{tabular}{cgggggggggggggg} 
				\toprule
				\rowcolor{white}
				Network's name & Sub-networks name & Trainable parameters & Layer configuration \\
				\toprule


 \rowcolor{lg}
 \texttt{NA}& -& 905931 & 100, 500, 800, 500, 100\\
\rowcolor{white}
 \texttt{NA ensemble} & Forward networks & 64531  $\times$ 10 & 100, 100, 200, 100, 100 \\
 \rowcolor{lg}
 \texttt{UANA}&  $\mu$ networks & $64531 \times 10$  & 100, 100, 200, 100, 100 \\
  \rowcolor{lg}
 &  $\sigma$ networks & $24231 \times 10$  & 100, 100, 100 \\
\rowcolor{white}
 \texttt{Tandem}&Forward network& 905931 & 100, 500, 800, 500, 100 \\
 \rowcolor{white}
&Inverse network& 117108 & 100, 250, 250, 100\\
\rowcolor{lg}
& $\mu$ networks& $64531 \times 10$  & 100, 100, 200, 100, 100\\
 \rowcolor{lg}
 \texttt{UATandem}& $\sigma$ networks & $24231  \times 10 $ & 100, 100, 100\\
 \rowcolor{lg}
&Inverse network& 117108 & 100, 250, 250, 100 \\

				\bottomrule
			\end{tabular}
		
		\label{tab:networks_param_specral_printing-config}
	\end{table}
	}
\newcommand{\SpectralPrintingInk}{
\begin{table*}
		\centering
		\caption{The distribution of ink densities ($\geq 0.4$) after the inversion of \texttt{spectral printer} using \texttt{UA-tandem}. Once we insert noise into LC channel or sample it sparsely, \texttt{Autoinverse} detects and avoids it. STD has been rounded to nearest integer.}
                \resizebox{\linewidth}{!}{%
			\begin{tabular}{cgggggggggggggg} 
				\toprule
				\rowcolor{white}
	
				Model & Data set & NFP error &
				\cellcolor{cyan!70}C & \cellcolor{magenta!70}M & \cellcolor{yellow!70}Y & \cellcolor{black!50}K &
				\cellcolor{cyan!20} LC & \cellcolor{magenta!20} LM &
				\cellcolor{gray!30} LK & \cellcolor{gray!20} LLK\\
				\toprule

\rowcolor{white}
  &Standard& $(2.62 \pm 0.488) \times 10^{-3}$ & $180 \pm 12$ & $48 \pm 2$ & $13 \pm 4$ & $3 \pm 1$ &\cellcolor{green!30} $174 \pm 8$ & $36 \pm 12$ & $0 \pm 0$ & $0 \pm 0$\\

 \rowcolor{white}
 \texttt{UA-tandem}&Sparse& $(2.34 \pm 0.097) \times 10^{-3}$ & $291 \pm 11$ & $43 \pm 0$ & $15 \pm 2$ & $3 \pm 1$ &\cellcolor{green!30} $\mathbf{0 \pm 0}$ & $89 \pm 16$ & $0 \pm 0$ & $0 \pm 0$\\

\rowcolor{white}
 &Noisy& $(5.16 \pm 0.423) \times 10^{-3}$ & $242 \pm 2$ & $63 \pm 0$ & $18 \pm 1$ & $1 \pm 0$ &\cellcolor{green!30} $\mathbf{0 \pm 0}$ & $64 \pm 32$ & $8 \pm 1$ & $29 \pm 2$\\

  \cmidrule(lr){2-11}
 
\rowcolor{white}
  &Standard& $(3.51 \pm 2.903) \times 10^{-2}$ & $159 \pm 9$ & $40 \pm 10$ & $0 \pm 0$ & $12 \pm 9$ &\cellcolor{green!30} $255 \pm 124$ & $187 \pm 256$ & $5 \pm 4$ & $1070 \pm 1514$\\

 \rowcolor{white}
 \texttt{tandem}&Sparse& $(2.18 \pm 0.840) \times 10^{-2}$ & $208 \pm 22$ & $36 \pm 3$ & $0 \pm 0$ & $17 \pm 7$ &\cellcolor{green!30} $52 \pm 30$ & $20 \pm 9$ & $14 \pm 19$ & $0 \pm 0$\\

\rowcolor{white}
 &Noisy& $(3.82 \pm 0.810) \times 10^{-2}$ & $174 \pm 29$ & $60 \pm 32$ & $39 \pm 42$ & $71 \pm 9$ &\cellcolor{green!30} $303 \pm 171$ & $192 \pm 239$ & $114 \pm 123$ & $62 \pm 74$\\
				\bottomrule
			\end{tabular}
		}

		\label{tab:ink_densities_distribution_(UA)tandem}
\end{table*} }

\newcommand{\SoftRobotLoss}{
\definecolor{lg}{gray}{0.92}
\definecolor{lgg}{gray}{0.975}
\newcolumntype{g}{>{\columncolor{lg}}c}
	\begin{table*}
		\centering
		\caption{[Continued] Training details of different neural surrogate models used in inverse methods for \texttt{soft robot}.}
                \small
			\begin{tabular}{cgggggggggggggg} 
				\toprule
				\rowcolor{white}
				Network's name & Sub-networks name  & Total training time (s) & Total inversion time (s) & Training loss\\
				\toprule


 \rowcolor{white}
 \texttt{NA}&-& $5428$ & 2250 & $2.63 \times 10^{-4}$\\
 \rowcolor{lg}
 \texttt{UANA}& $\mu$ networks & $16940$ & 2950 & $2.39 \times 10^{-5}$\\
  \rowcolor{lg}
 &  $\sigma$ networks & $16290$ &  & \\
\rowcolor{white}
 \texttt{Tandem}& Forward network& $5428$ & $2.90 \times 10^{-1}$ & $2.63 \times 10^{-4}$\\
 \rowcolor{white}
&Inverse network& $1740$ &  &  \\
\rowcolor{lg}
& $\mu$ networks  & $16940$ &  & \\
 \rowcolor{lg}
 \texttt{UATandem}& $\sigma$ networks & $16290$ & $1.65 \times 10^{-1}$ & $2.39 \times 10^{-5}$\\
 \rowcolor{lg}
&Inverse network& $9458$ &  & \\

				\bottomrule
			\end{tabular}

		\label{tab:networks-param-soft-robot-loss}
	\end{table*}
}

\newcommand{\SoftRobotConfig}{
\definecolor{lg}{gray}{0.92}
\definecolor{lgg}{gray}{0.975}
\newcolumntype{g}{>{\columncolor{lg}}c}
	\begin{table*}
		\centering
		\caption{Training details of different neural surrogate models used in inverse methods for \texttt{soft robot}.}
                \small
			\begin{tabular}{cgggggggggggggg} 
				\toprule
				\rowcolor{white}
				Network's name & Sub-networks name & Trainable parameters & Layer configuration \\
				\toprule


 \rowcolor{white}
 \texttt{NA}&-& 45506206 & 2000, 5000, 5000, 2000 \\
 \rowcolor{lg}
 \texttt{UANA}&  $\mu$ networks & $3227606 \times 10$ & 300, 1500, 1500, 300 \\
  \rowcolor{lg}
 &  $\sigma$ networks & $45106 \times 10$  & 100, 100, 100 \\
\rowcolor{white}
 \texttt{Tandem} & Forward network & 45506206 & 2000, 5000, 5000, 2000 \\
 \rowcolor{white}
& Inverse network & 3227440 & 300, 1500, 1500, 300 \\
\rowcolor{lg}
& $\mu$ networks & $3227606 \times 10$ & 300, 1500, 1500, 300 \\
 \rowcolor{lg}
 \texttt{UATandem}& $\sigma$ networks & $45106 \times 10$  & 100, 100, 100 \\
 \rowcolor{lg}
&Inverse network& 3227440 & 300, 1500, 1500, 300 \\

				\bottomrule
			\end{tabular}

		\label{tab:networks-param-soft-robot-config}
	\end{table*}
}

\newcommand{\ValidDesignSpectralPrint}{
\begin{figure}[t] 
\centering
\includegraphics[width = 1\linewidth]{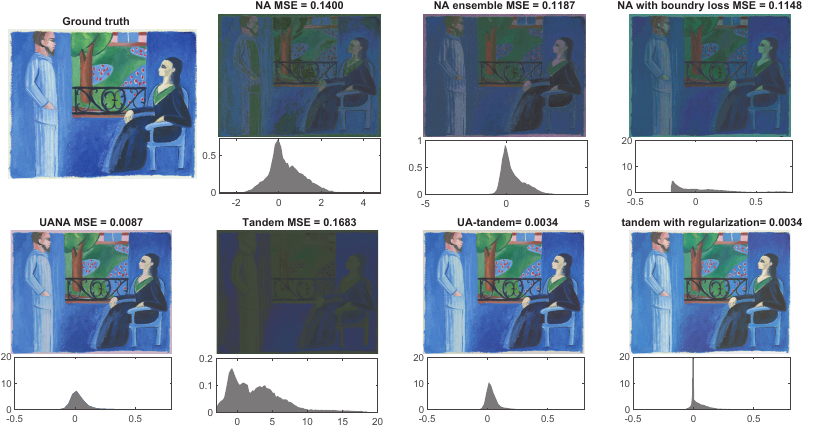}
\caption{Spectral reproduction of a ground-truth painting using different inverse methods (\texttt{spectral printer}). Apart form the reproduction quality, we show the distribution of the computed \textit{designs}, i.e., ink densities. Note that the feasible ink density range is $[0,1]$.}
\label{fig:valid_design_suplimenmtary}
\end{figure}
}
\newcommand{\SoftRobotLossTarget}{
\begin{equation}
\begin{split}
\mathcal{L}(\mathbf{x}^{in}, \mathbf{x}^{out}):=\left\| \mathbf{x}_{i}^{out} -\mathbf{t}\right\|_{1}+\lambda \cdot \mathcal{R}(\mathbf{x}^{in})\\
i \in \left[123, \ 124  \right],
\label{eq:SoftRobotObj_NA}
\end{split}
\end{equation}}
\newcommand{\SmoothnessLoss}{
\begin{equation}
\mathcal{R}(\mathbf{x}^{in}):= \sum_{1<i<n, i \neq n / 2, \atop i \neq n / 2+1}
\left( \frac{\mathbf{x}^{in}_{i+1}-\mathbf{x}^{in}_{i}}{2}-\frac{\mathbf{x}^{in}_{i}-\mathbf{x}^{in}_{i-1}}{2}\right)^{2}.\\
\label{eq:SoftRobot_smoothness_NA}
\end{equation}}
%
\newcommand{\EpistemicAleatoryPlot}{
\begin{figure}
\centering     
\subfigure[]{
\label{fig:SoftRobotDistribution_UANA_aleatory}
\includegraphics[width=0.48\textwidth]{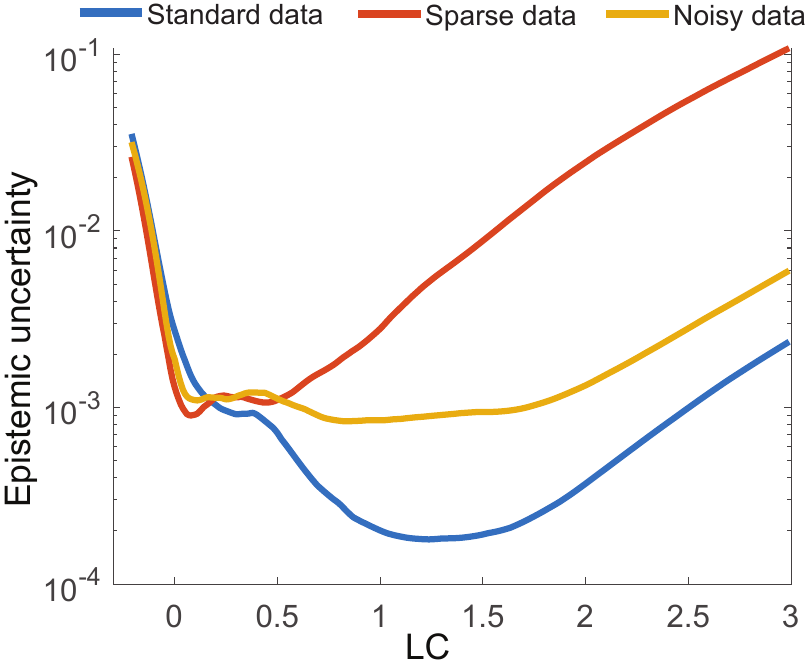}
\label{fig:epistemic_uncertainty_plot}}
\hfill
\subfigure[]{
\label{fig:SoftRobotDistribution_UANA_epistemic}
\includegraphics[width=0.48\textwidth]{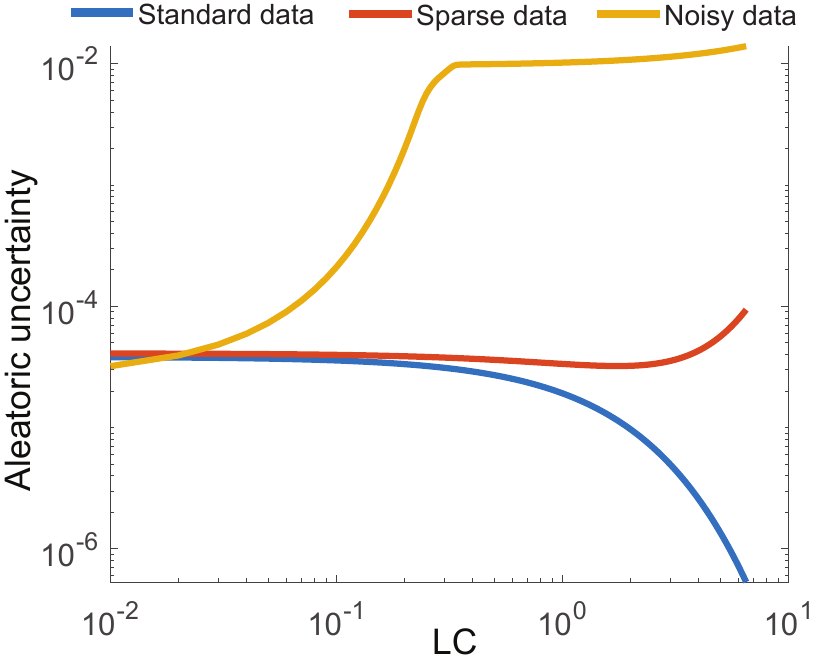}
\label{fig:aleatory_uncertainty_plot}}
\caption{{The landscape of the aleatory and epistemic uncertainty.}}
\end{figure}
}
%
\newcommand{\AlphaBetaPlot}{
\begin{figure}
\centering     
\subfigure[]{
\label{fig:alpha}
\includegraphics[width=0.48\textwidth]{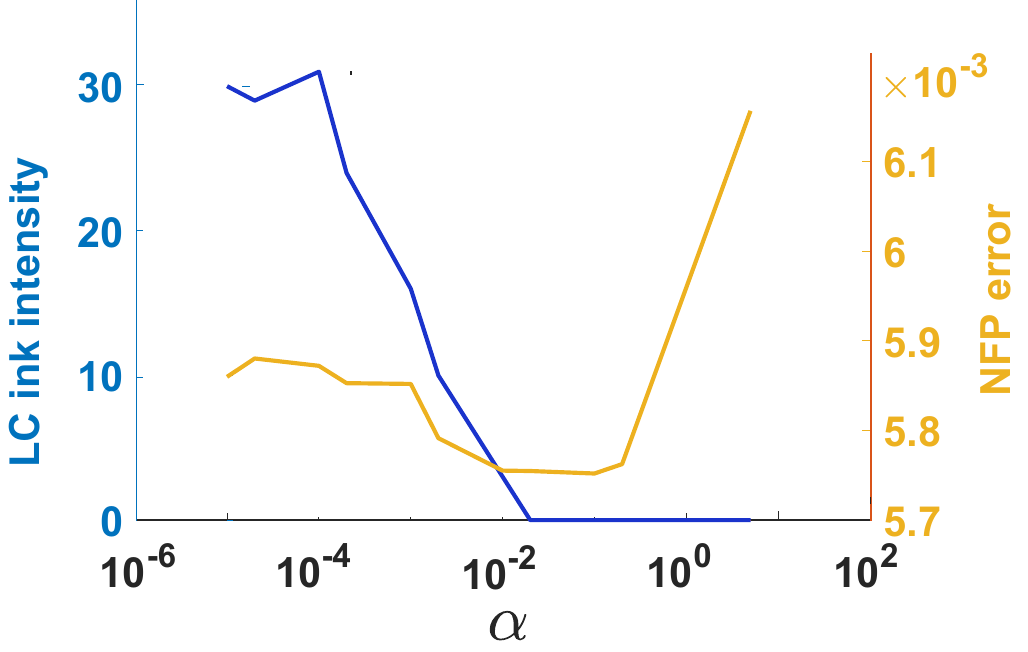}}
\hfill
\subfigure[]{
\label{fig:beta}
\includegraphics[width=0.48\textwidth]{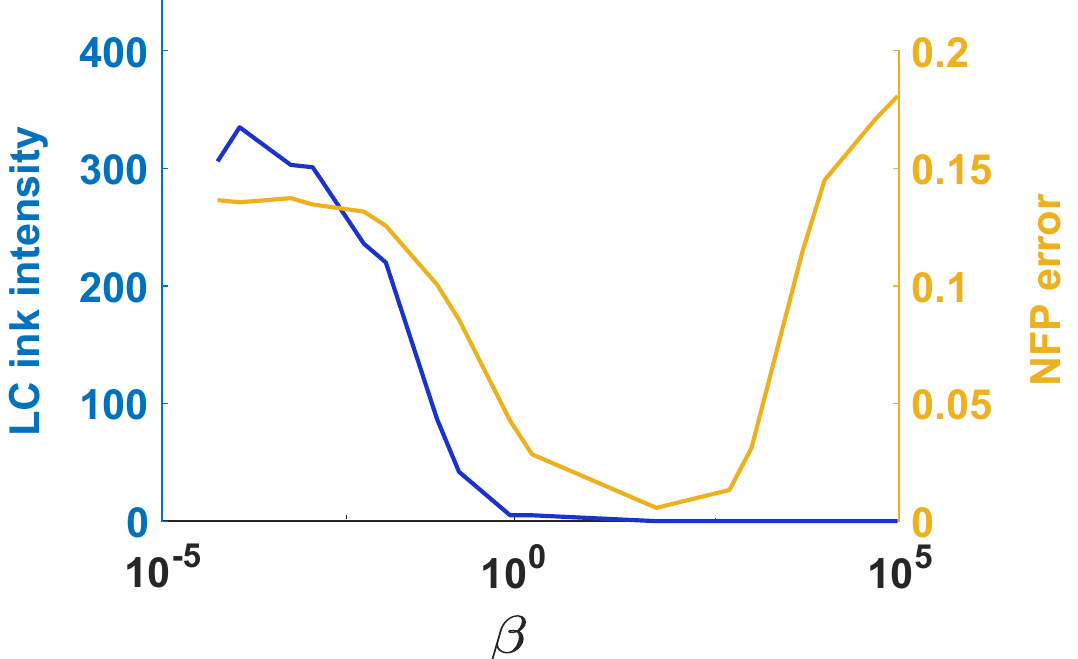}}
\caption{Stability of \texttt{Autoinverse} within a wide range of its hyperparameters $\alpha$ and $\beta$.}
\end{figure}
}

\newcommand{\WrongInit}{
\begin{figure}[t] 
\centering
\includegraphics[width = 1\linewidth]{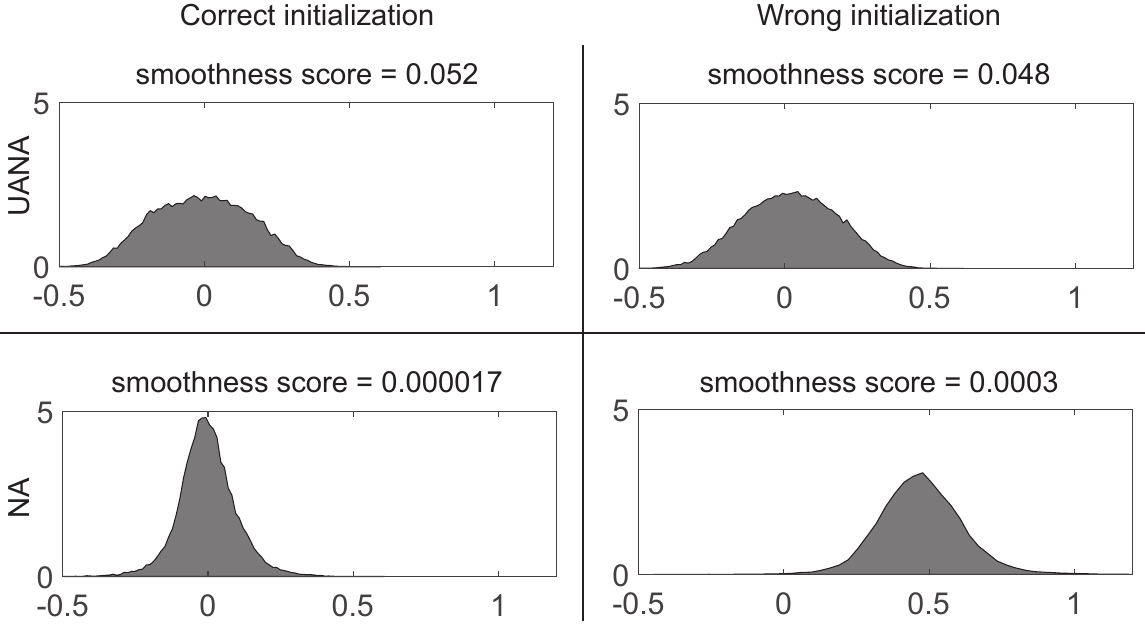}
\caption{The effect of wrong initialization for inversion of \texttt{soft robot} using \texttt{NA} with regularization and two different initialization.
Note the robustness of \texttt{UANA} without any form of regularization.}
\label{fig:soft_robot_wrong_init_sm}
\end{figure}}

\newcommand{\SoftRobotDistributionUANA}{
\begin{figure}
\centering     
\subfigure[16th edge is noisy in the positive actuation range.]{
\label{fig:SoftRobotDistribution_UANA_aleatory}
\includegraphics[width=0.48\textwidth]{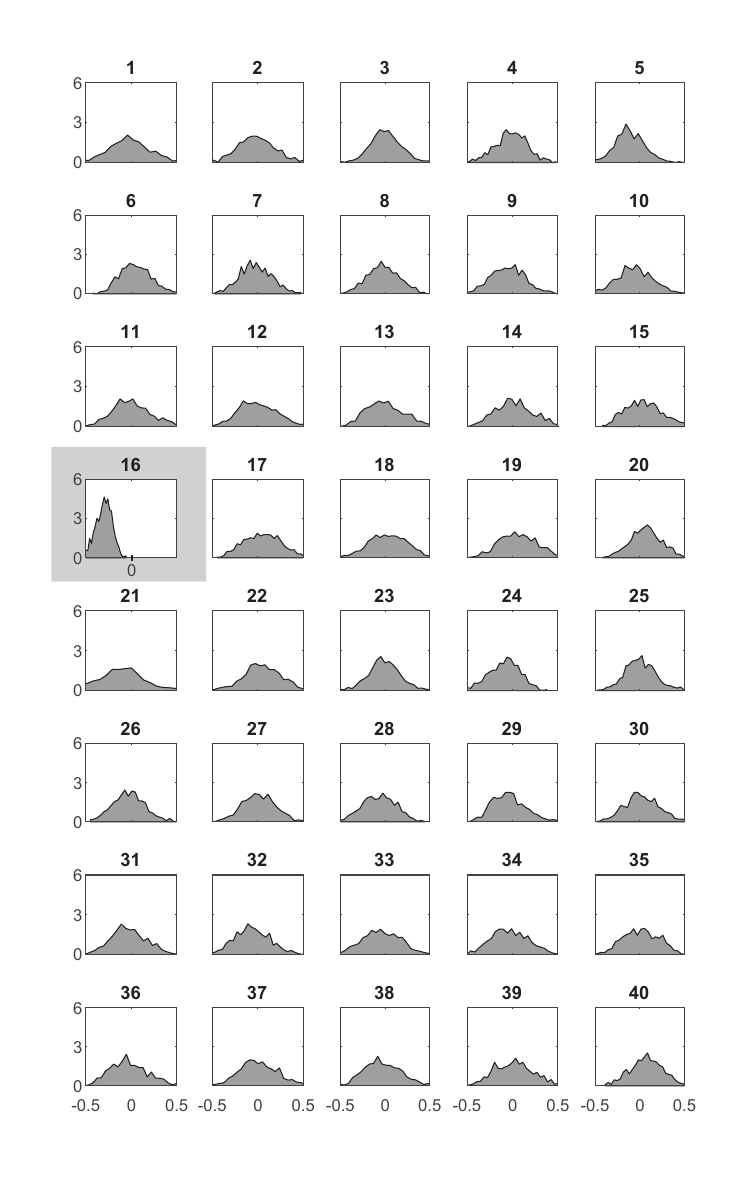}}
\hfill
\subfigure[{The training data does not contain any samples with positive actuation of the 16th edge.}]{
\label{fig:SoftRobotDistribution_UANA_epistemic}
\includegraphics[width=0.48\textwidth]{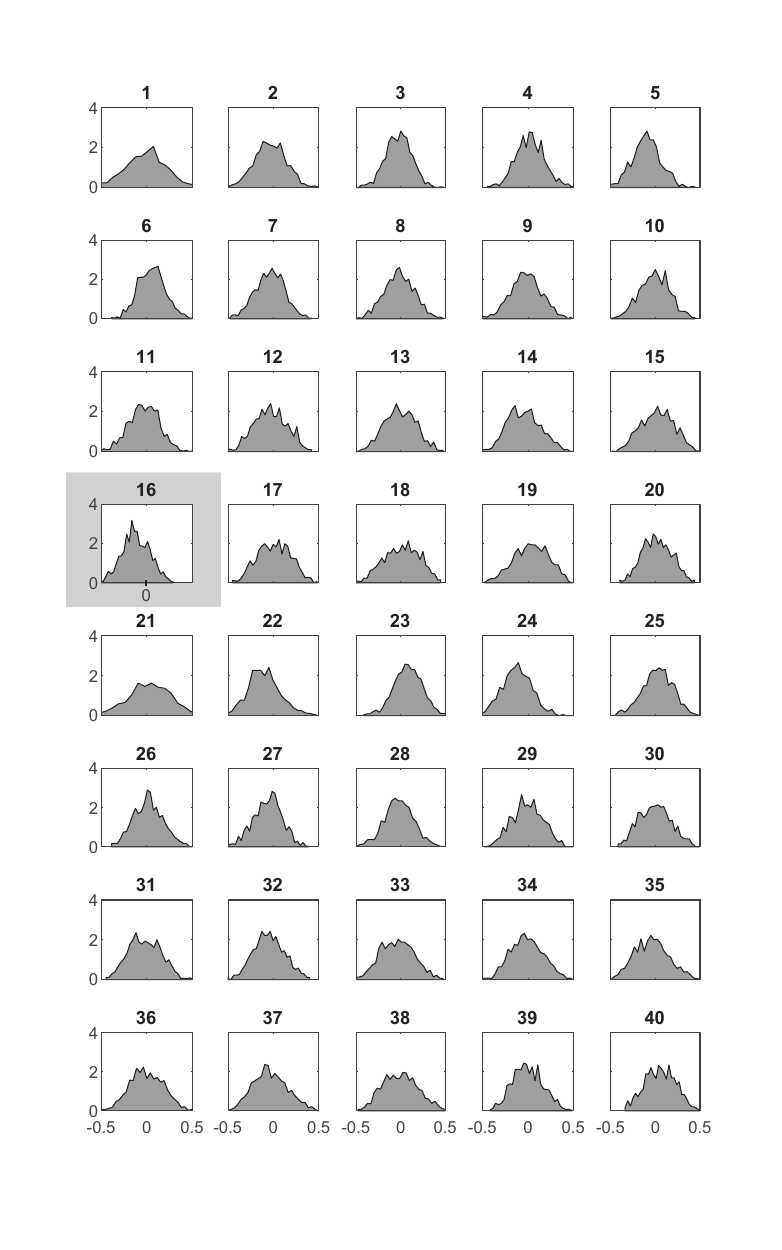}}
\caption{The actuation distribution of all the edges for inversion on both noisy and sparse data via \texttt{UANA}.}
\label{fig:SoftRobotDistribution_UANA_label}
\end{figure}
}
\newcommand{\NAEnsembleArch}{
\begin{figure}[b]
\centering
\includegraphics[width = 0.6\linewidth]{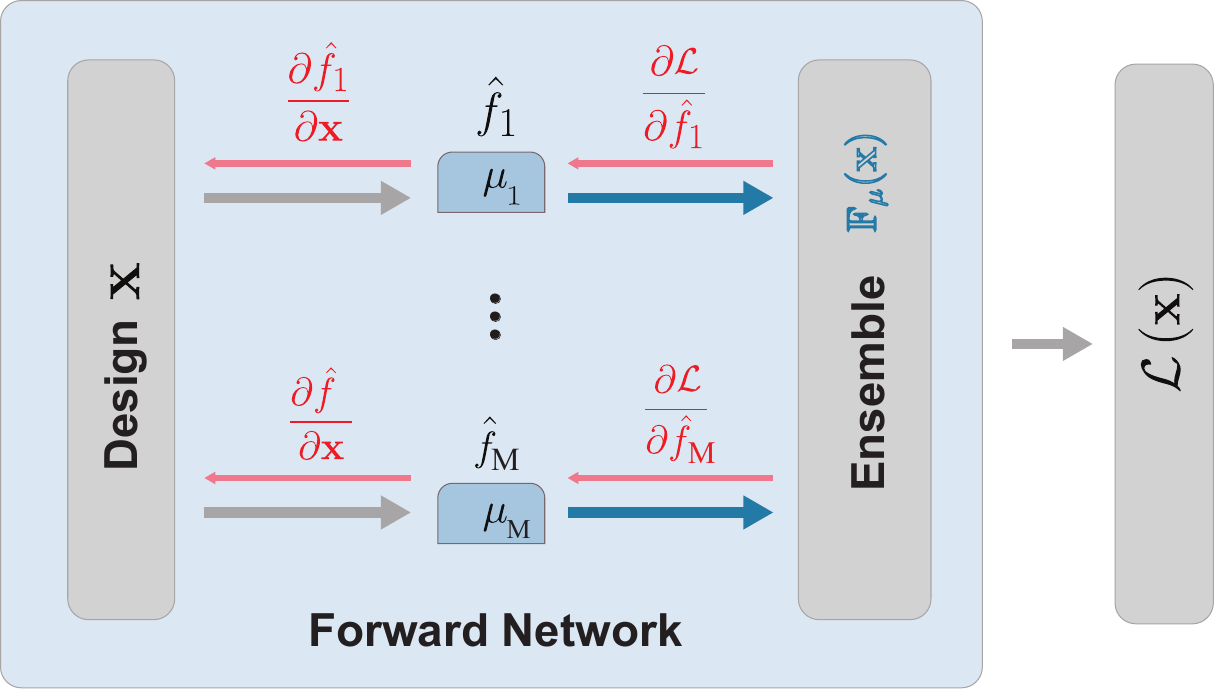}
\caption{NA ensemble architecture}
\label{fig:NAEnsembleArchitecture}
\end{figure}}
\newcommand{\SoftRobotDistributionUAT}{
\begin{figure}
\centering     
\subfigure[16th edge is noisy in the positive actuation range.]{
\label{fig:SoftRobotDistribution_aleatory_UAT}
\includegraphics[width=0.48\textwidth]{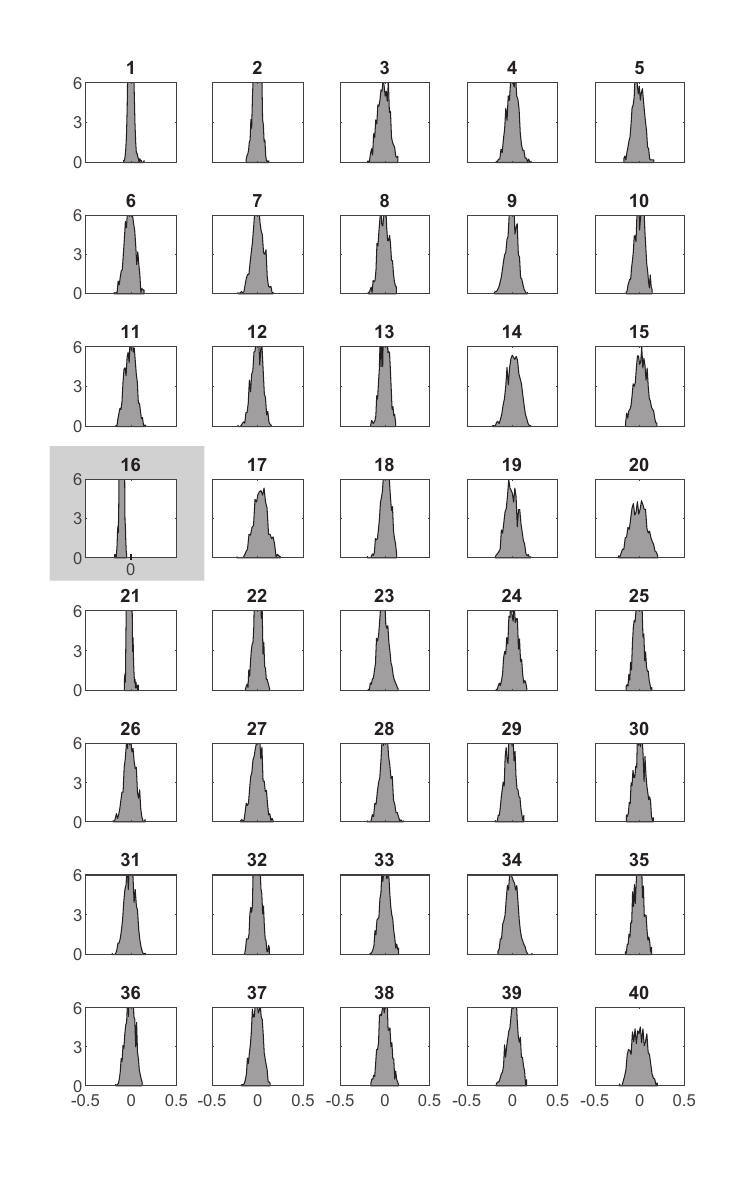}}
\hfill
\subfigure[{The training data does not contain any samples with positive actuation of the 16th edge.}]{
\label{fig:SoftRobotDistribution_epistemic_UAT}
\includegraphics[width=0.48\textwidth]{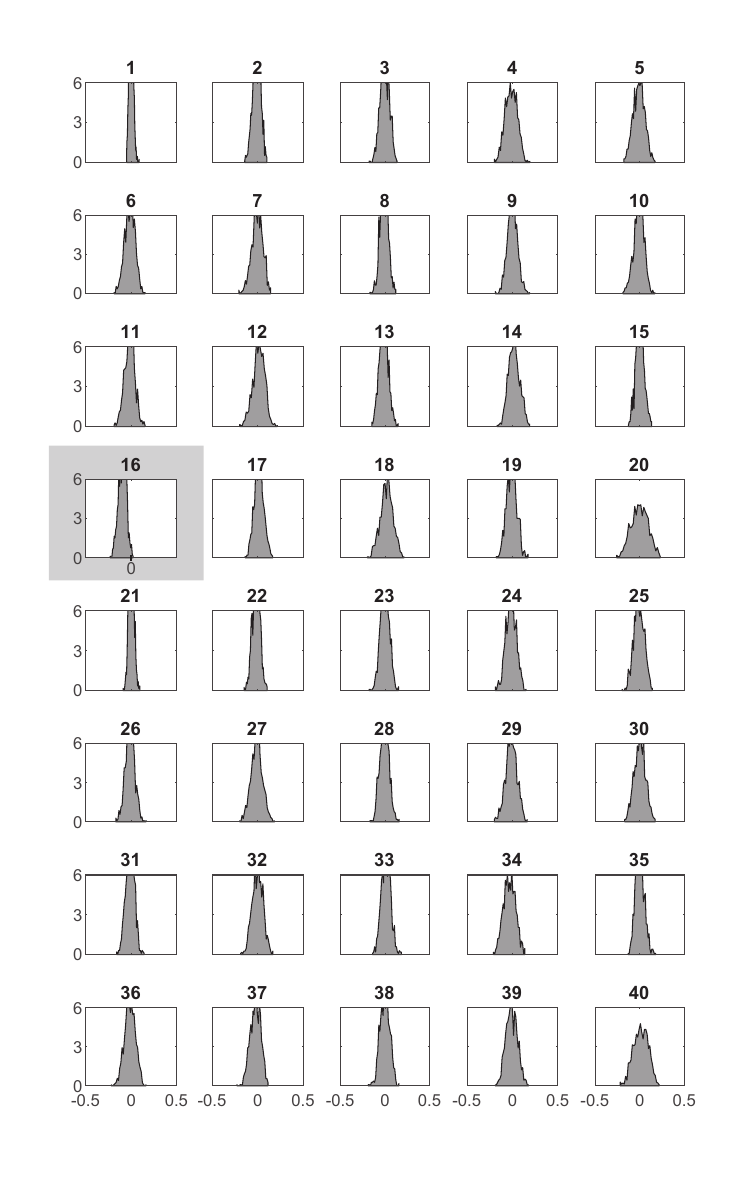}}
\caption{The actuation distribution of all the edges for inversion on both noisy and sparse data via \texttt{UA-tandem}.}
\label{fig:SoftRobotDistribution_UAT_label}
\end{figure}
}

\newcommand{\Paretofront}{
\begin{figure}[t] 
\centering
\includegraphics[width = 1\linewidth]{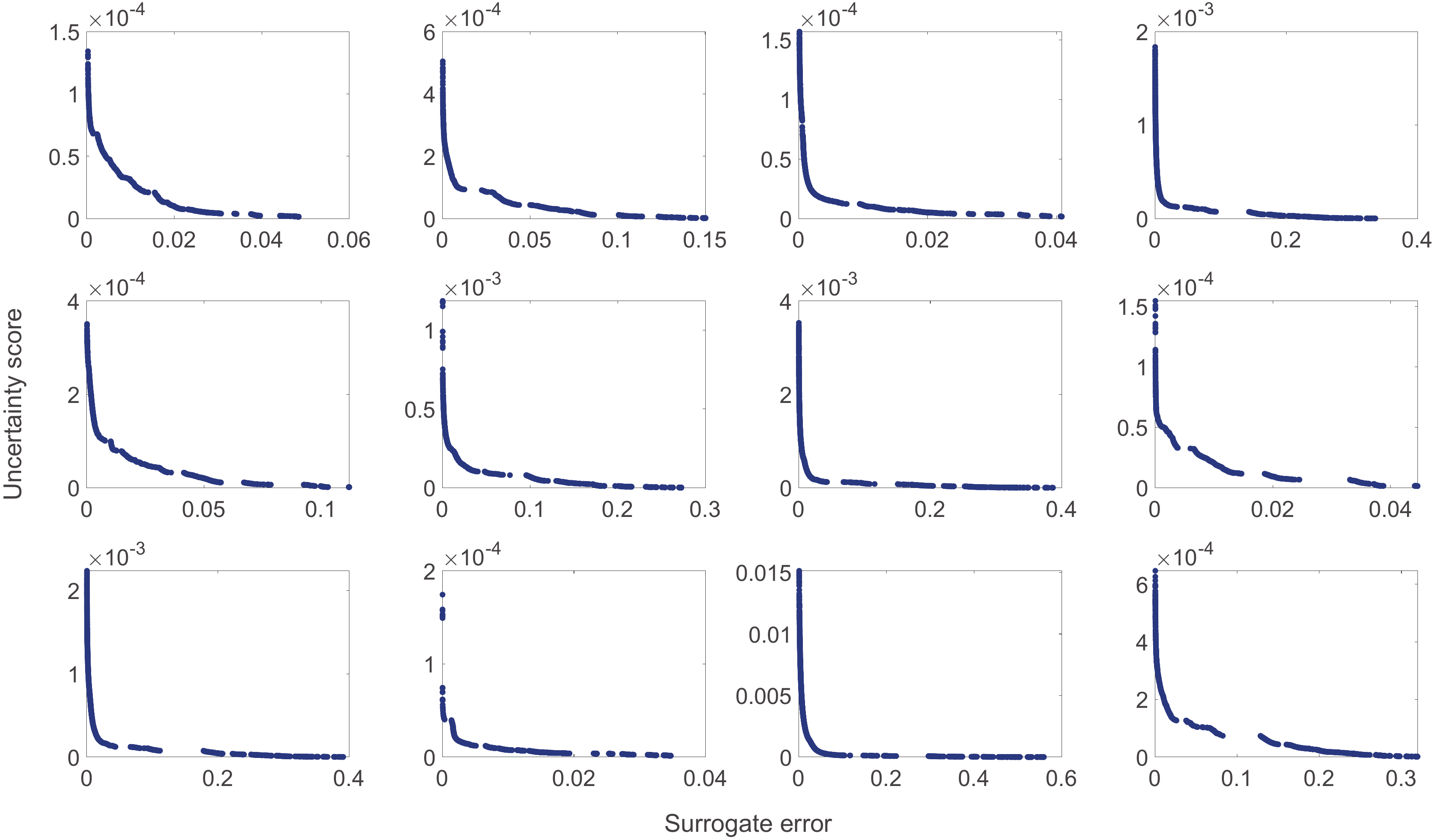}
\caption{Pareto front for 12 randomly chosen targets.}
\label{fig:pareto}
\end{figure}}
\newcommand{\numberofnetworks}{
\begin{figure}
\centering     
\subfigure[The effect of increasing the number of networks in the ensemble trained on the noisy data.]{
\label{fig:numberofnetworks_noisy}
\includegraphics[width=0.48\textwidth]{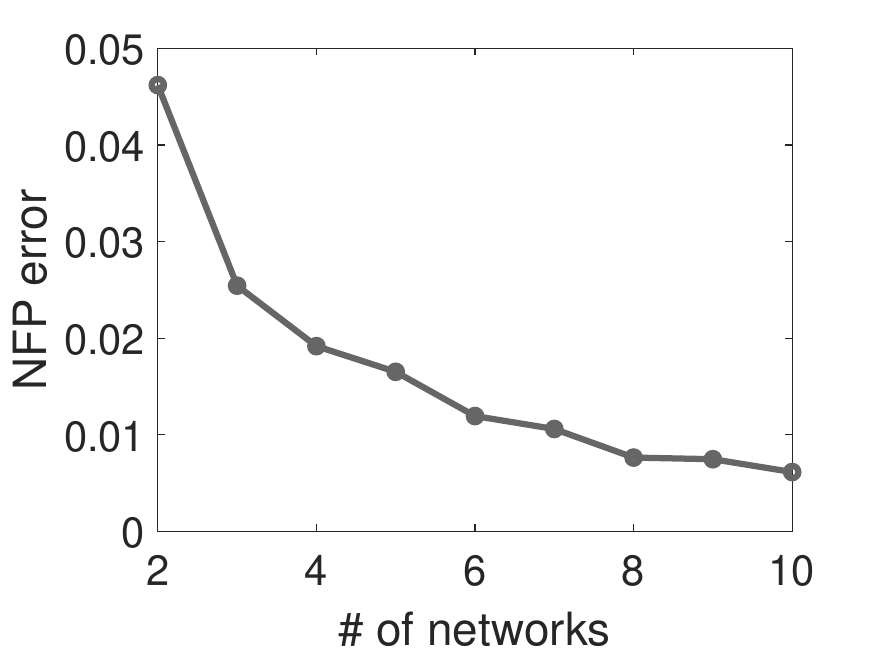}}
\hfill
\subfigure[The effect of increasing the number of networks in the ensemble trained on the sparse data.]{
\label{fig:numberofnetworks_sparse}
\includegraphics[width=0.48\textwidth]{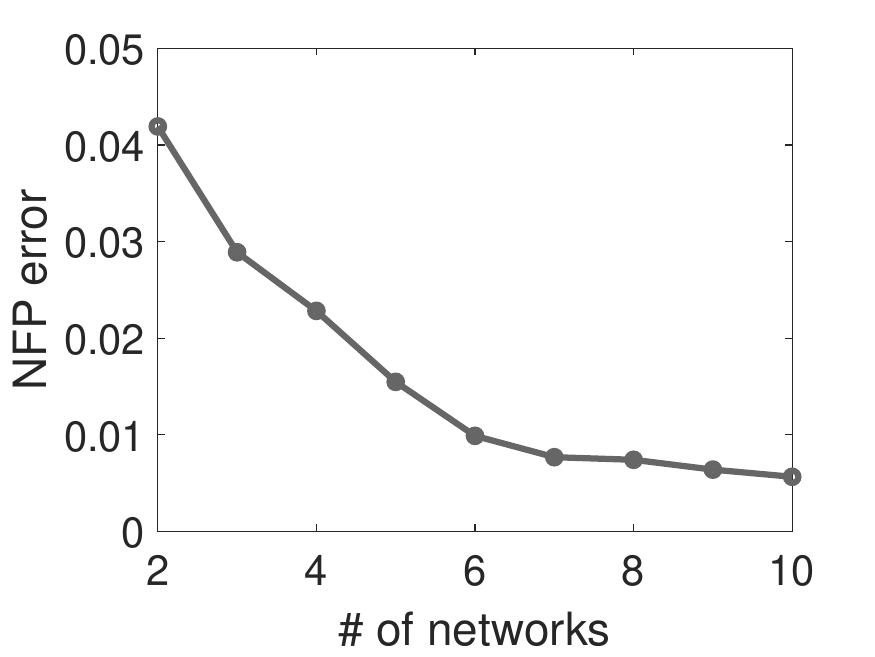}}
\caption{Evaluating the inversion performance using Deep Ensembles with different number of sub-networks. }
\label{fig:numberofnetworks}
\end{figure}
}

\section{Standard error calculation, tuning, and implementation details}\label{sec:ste_tuning_imp_details}
In this section we provide more details on different inverse methods we use in the paper. 
As many of the inverse methods we use in the paper have a stochastic component, we perform all following experiments 3 times and report the standard {error}.
Since we have a budget of 5 runs for tuning the hyperparameters of \texttt{Autoinverse} methods, we allocate similar or higher resources for other methods in order to make the comparisons fair. 

\paragraph{\texttt{NA}}
To avoid local minima, we run 50 \textit{solves} of inversion with random initialization (each having up to 2000 iterations).
As an alternative to tuning, we run \texttt{NA} 5 times and report the best NFP error.
To accelerate the inversion in this paper we perform batch optimization.
Thus, we can increase the number of target samples (up to GPU memory limit) without impacting the inversion time noticeably.

\paragraph{\texttt{NA ensemble}} 
The configurations for \texttt{NA ensemble} is similar to \texttt{NA} except for the forward model.
Unlike \texttt{NA}, we have an ensemble of networks ($\mathbb{F}_{\mathbf{\mu}}$) that comprises the forward model of \texttt{NA ensemble}:
\begin{equation}
\mathbb{F}_{\mathbf{\mu}}(\mathbf{x}):= \frac{1}{M} \ \sum_{m} \ \mu_{m}(\mathbf{x}) \label{eq:F_mu}.
\end{equation}
Unlike \texttt{UANA}, single networks in the ensemble are incapable of predicting uncertainty. The cost function for \texttt{NA ensemble} is therefore defined as:
\begin{equation}
    \mathcal{L}^{NA_{en}}(\mathbf{x}) := \ \arg\min_{x} \left\| \mathbb{F}_{\mathbf{\mu}}(\mathbf{x}) -  \ \mathbf{y^{*}} \right\|_{2}^{2}
\label{eq:NAensemble_loss}
\end{equation}
Similar to \texttt{UANA}, the back propagation is based on an ensemble of gradients coming from all the single networks in the ensemble (Figure~\ref{fig:NAEnsembleArchitecture}).
\begin{equation}
    \mathbf{x}^{z} = \mathbf{x}^{z-1} - \delta\sum_{m=1}^{M}(\frac{\partial \mathcal{L}^{NA_{en}}}{\partial \hat{f}_m} \times \frac{\partial \hat{f}_m}{\partial \mathbf{x}}) 
    \label{eq:NA_ensemble_grad}
\end{equation}
Similar to \texttt{NA}, we run \texttt{NA ensemble} 5 times and pick the best results with respect to the NFP error.
\NAEnsembleArch

\paragraph{\texttt{tandem}}
Since \texttt{tandem} does not explicitly possess any hyperparameter, instead of tuning, we evaluate 5 pre-trained inverse models (each initialized differently) and select the one that has the best NFP error {on 10\% of the target data}.
We then perform the inversion on the remaining target data using the best-performing inverse model.
%

\paragraph{\texttt{UANA} and \texttt{UA-tandem}}
In Section~{4.5} of paper we showed how using a diverse range of activation functions in the ensemble network improves the quality of epistemic uncertainty and consequently the results of the inversion.
For training the surrogates used in \texttt{UANA}, \texttt{UA tandem} and \texttt{NA ensemble} we use 10 networks in the ensemble with the following activations $\mathbb{F}_{\mu}$:

\begin{itemize}
    \item Tanh $\times 2$
    \item ReLU $\times 2$
    \item CELU $\times 2$
    \item LeakyReLU $\times 2$
    \item ELU
    \item Hardswish
\end{itemize}
We use ReLU activation functions for all other methods.

\paragraph{\texttt{INN}}
For each one of the 1000 target performances we randomly sample the latent space of the INN architecture \cite{ren2020benchmarking} 1024 times.  
Thus, we end up with 1024 designs (for a single target) and evaluate all designs on the NFP and report the best error as the NFP error.
For the surrogate error, we report the average forward loss.

\paragraph{\texttt{MINI}}
\texttt{MINI} is based on a mixed-integer optimization which is capable of finding the globally optimum solution \cite{ansari2021mixed}.
This method is deterministic and every run returns the same solution, as a result we do not report the standard error for \texttt{MINI}.

\paragraph{Hardware}
To have a fair comparison we run all the methods on the same GPU machine for evaluating time performances. 
We used an NVIDIA TITAN X GPU for time evaluation.
For other evaluations we used a GPU cluster.
Training the forward models is trivially parallelizable.
Moreover, we can parallelize 50 iterations of \texttt{NA} and \texttt{UANA} and aggregate the data in a post-processing step and choose the best results \textit{based on the surrogate error}.
Nevertheless, we are reporting our computation time assuming {both training and inversion are performed} serially on a single GPU.
%
\section{Training details for neural surrogate models in \texttt{multi-joint robot} (Table~1 in paper)}
\MultiJointRobotConfig
\MultiJointRobotTrainingLoss
In Table~\ref{tab:networks-param-robotic-arm-config} and Table~\ref{tab:networks-param-robotic-arm-loss}, we can compare the capacity, training time, and accuracy of the neural network surrogates used for the inversion of \texttt{multi-joint robot}.
We keep similar training capacity for all methods except \texttt{MINI}.
\texttt{MINI} uses a combinatorial optimization and is not scalable to large networks \cite{ansari2021mixed}.
We trained a smaller surrogate for \texttt{MINI} but at the same time we monitored its training loss to lie within a reasonable range.
%
\section{Details for `Neural inversion in the presence of imperfect data'}\label{sec:epistemic_aleatoric_uncertainty_sm}
\subsection{Counterpart results of \texttt{tandem} and \texttt{UA-tandem} for \texttt{spectral printer}}
\SpectralPrintingInk
Table~\ref{tab:ink_densities_distribution_(UA)tandem} presents the results of \texttt{spectral printer}, similar to Section~{4.3} (Table~{2}) of the paper but comparing \texttt{tandem} and \texttt{UA-tandem}.
As evident from the table, we obtain similar performance gain when augmenting \texttt{tandem} with uncertainty awareness.
Similar to \texttt{UANA}, \texttt{UA-tandem} has completely avoided the LC channel.

\subsection{Soft robot actuation distribution}
Figure~{2} in the paper showed the actuation distribution of only 8 soft robot edges (Section~{4.3} in paper). In Figure~\ref{fig:SoftRobotDistribution_UANA_label} we show the distribution of actuation for \textit{all} edges computed by \texttt{UANA} on surrogates learned using partially noisy and partially sparse data.
\SoftRobotDistributionUANA
We repeat this experiment using \texttt{UA-tandem} to emphasize on the generality of \texttt{Autoinverse} (Figure~\ref{fig:SoftRobotDistribution_UAT_label}).
\SoftRobotDistributionUAT

\subsection{Training details of surrogates used for \texttt{spectral printer} and \texttt{soft robot}}
\SpectralPrintingConfig
\SpectralPrintingLoss
\SoftRobotConfig
\SoftRobotLoss
Tables \ref{tab:networks_param_specral_printing-config} and \ref{tab:networks_param_specral_printing-loss}  show the layer configuration, number of trainable parameters, the training and inversion time, and the training loss of different surrogate models used in \texttt{spectral printer}.
Tables \ref{tab:networks-param-soft-robot-config} and \ref{tab:networks-param-soft-robot-loss}  show the layer configuration, number of trainable parameters, the training and inversion time, and the training loss of different surrogate models used in \texttt{soft robot}.
From the tables we can observe that the training capacity of different surrogates models is comparable.
Furthermore, the training accuracy of all models is similar.

\subsection{Epistemic and aleatoric loss behaviour for spectral printing experiment}\label{sec:SM_epistemic_aleatoric}
\paragraph{Epistemic uncertainty}
    The key to handling design feasibility is the epistemic uncertainty (Equation 3(c) paper).
    We know that the scarcity of data results in higher epistemic uncertainty \cite{kendall2017uncertainties}.
    On the other hand, by definition, we do not have any infeasible or out-of-range data points in our dataset.
    Hence, if we query a network with an infeasible or out-of-range input, we will get high uncertainty for the prediction.
    We use this trend to avoid such samples in the inversion.
    Figure~\ref{fig:epistemic_uncertainty_plot} shows the trend of the epistemic uncertainty values for \texttt{spectral printer} (Section~{4.3}, Table~{2} in paper).
    For that experiment, we ran the inversion using \texttt{UANA} on 3 different datasets: standard, noisy, and sparse and observed how the problematic ink channel (LC) is avoided. 
    Figure~\ref{fig:epistemic_uncertainty_plot} demonstrates \textit{why} that ink channel is avoided. 
    In Figure \ref{fig:epistemic_uncertainty_plot}, we set all ink channels except LC to 0 while increasing the values of LC ink density from 0 (the x-axis of the plot).

    As expected, for all three datasets moving away from the feasible region (between 0 and 1) increases the epistemic uncertainty.
    When trained for the sparse data (red curve), where the LC channel has not been sampled after 0.4, the epistemic uncertainty starts to increase earlier. 
    Outside the feasible region, each network in the ensemble has to extrapolate as it has not been trained in those regions. 
    Consequently, the predictions of ensemble networks diverge.
    The divergence of the networks increases the epistemic uncertainty and, during the inversion, the uncertainty aware methods can reject solutions in these regions.
\paragraph{Aleatoric uncertainty}
    Figure \ref{fig:aleatory_uncertainty_plot} demonstrates the behavior of the aleatoric uncertainty of the surrogate used for the same experiment (\texttt{UANA} on \texttt{spectral printer}).  
    Similarly, to generate the plots, we set all the ink channels to 0 and change the values of the Light Cyan ink densities.
    As evident from Figure \ref{fig:aleatory_uncertainty_plot}, the level of uncertainty increases significantly for the noisy dataset, while for sparse and standard datasets aleatoric uncertainty is at least two orders of magnitude smaller.
    The increase of aleatoric uncertainty for the noisy data helps \texttt{UANA} avoid any samples from those regions (Tables~{2} in the paper and Table~\ref{tab:ink_densities_distribution_(UA)tandem}).

\EpistemicAleatoryPlot

%
\section{Details for `Autoinverse brings AutoML to neural inversion'}\label{sec:SM_automl_neural_inversion}
\paragraph{Spectral printer}
\ValidDesignSpectralPrint
Complementary to Figure~{4} of the paper, in Figure \ref{fig:valid_design_suplimenmtary} we compare the inversion performance using a diverse range of inverse methods.
Here we clearly see that basic methods, such as \texttt{NA} and \texttt{tandem} fail spectacularly in computing feasible designs.

{Boundary loss} is a semi-generic regularization, suitable for handling box constraints \cite{ren2020benchmarking}: 
\begin{equation}
\mathcal{L}_{b n d}=\operatorname{Re} L U\left(\left|x-\mu_{x}\right|-\frac{1}{2} R_{x}\right)
\end{equation}%
where $R_{x}$ is the value range of design samples in the dataset and $\mu_{x}$ is their average.
Incorporating the boundary loss in \texttt{NA} in Figure \ref{fig:valid_design_suplimenmtary} results in an improvement in the distribution of the ink intensities. 
However, the results are still far from acceptable.

Note how applying hand-crafted regularization (ink intensity regularization) \cite{shi2018deep} improves the quality of \texttt{tandem} significantly (\texttt{tandem with regularization} in Figure \ref{fig:valid_design_suplimenmtary}). %
\texttt{UANA} and \texttt{UA-tandem} however perform comparably without any regularization.
\paragraph{Soft robot objective for the Neural Adjoint method}\label{sec:NA_softrobot_appendix}
The objective function for \texttt{soft robot} inversion comprises of two terms, one is responsible for bringing the tip of the robot to the target and the other one ($\mathcal{R}(\mathbf{x}^{in})$) guarantees the deformations to remain physical~\cite{sun2021amortized}.
\begin{equation}
\begin{split}
\mathcal{L}(\mathbf{x}^{in}):=\left\| \mathbf{x}_{i}^{out} -\mathbf{t}\right\|_{1}+\lambda \cdot \mathcal{R}(\mathbf{x}^{in})\\
i \in \left[123, \ 124  \right],
\label{eq:SoftRobotObj_NA}
\end{split}
\end{equation}
\begin{equation}
\mathcal{R}(\mathbf{x}^{in}):= \sum_{1<i<n, i \neq n / 2, \atop i \neq n / 2+1}
\left( \frac{\mathbf{x}^{in}_{i+1}-\mathbf{x}^{in}_{i}}{2}-\frac{\mathbf{x}^{in}_{i}-\mathbf{x}^{in}_{i-1}}{2}\right)^{2}.
\label{eq:SoftRobot_smoothness_NA}
\end{equation}
where $\mathbf{t}$ represents the target location and $\mathbf{x}_{i}^{out}$ represents the position of all 206 vertices of soft robot, among which $i \in \left[123, \ 124  \right]$ represent the position of its tip.
Also, $\lambda$ adjusts the importance of the smoothness term and $\mathcal{R}(\mathbf{x}^{in})$ regulates the actuation of the flexible edges ($\mathbf{x}^{in}$ ) to insure that the deformation of the robot is physical.

\paragraph{Sensitivity to initialization (\texttt{soft robot})}
In Section~{4.4} of the paper, we learned how hand-crafted regularization improves the quality of the designs.
Despite having regularization, \texttt{soft robot} inversion using \texttt{NA} fails when initialized with values far from feasible region.
This is evident from both the irregular robot shapes (Figure~3(b) in the paper) and the distribution of the actuation (Figure~\ref{fig:soft_robot_wrong_init_sm}) computed using \textbf{regularized} \texttt{NA} but with \textit{wrong} initialization. 
At the same time, \texttt{UANA} without any regularization and with a wrong initialization produces plausible robot shapes (Figure~3(b) in the paper) and actuation distribution (centered around 0). 
Interestingly, the smoothness score for clearly failed shapes obtained from \texttt{NA} with regularization, are for both initializations very small (Figure~\ref{fig:soft_robot_wrong_init_sm}).
This shows how even regularization can be multi-modal and fall into a wrong local minima and generate designs with nonphysical shapes (Figure 3(b)).
\WrongInit
\section{Details for `Ablation studies'}\label{sec:Abelation_details}
As discussed in Section~4.5 of the paper, the power of \texttt{UANA} lies in its uncertainty awareness and not the ensembling process.
We include the result of \texttt{NA ensemble} for paitning reproduction in Figure~\ref{fig:valid_design_suplimenmtary}. While we observe a marginal improvement of \texttt{NA ensemble} over \texttt{NA}, it is clearly outperformed by \texttt{UANA}.

\paragraph{Sensitivity to uncertainty weights}
\texttt{Autoinverse} is extremely stable when tuning its hyperparameters, i.e., uncertainty weights ($\alpha$ and $\beta$ in Equation~{3} in paper). This ensures {that} a light hyperparameter tuning is enough for obtaining reasonable results. 
We evaluate this behavior by using a wide range of $\alpha$ and $\beta$ values spanning over 5 orders of magnitude.
This ablation is performed using \texttt{UANA}.
In the ablation of aleatoric ($\alpha$) and epistemic ($\beta$) weight, we have used the noisy and epistemic data of \texttt{spectral printer}, respectively.
When evaluating $\alpha$ we keep $\beta$ fixed at the tuned value.
Alternatively for the evaluation of $\beta$, $\alpha$ is constant at the tuned value.
\AlphaBetaPlot

The most interesting fact about Figures \ref{fig:alpha} and \ref{fig:beta}, is the correlation of the $\alpha$ and $\beta$ with the NFP error.
This correlation means that adjusting the importance of these weights on the surrogate model directly improves the quality of the inversion in reality.
Interestingly, we can observe the robustness of \texttt{Autoinverse} against the variation of $\alpha$ and $\beta$, such that for a range of around 3 orders of magnitude the NFP loss remains stable around a desirable value.
The larger the weights, \texttt{Autoinverse} chooses less and less samples in the problematic regions (LC channel density larger than 0.4). This trend continues with very large uncertainty weights. 
However, these weights cannot be indefinitely increased as the MSE term of the objective ({Equations~6 and 10 in the paper}) will be undermined and inversion's NFP error increases.
%
\section{Pareto front of accuracy versus uncertainty}\label{sec:prato_front}
\Paretofront
We calculate the Pareto front for 12 randomly chosen targets from the spectral printer experiment with the standard dataset (Figure \ref{fig:pareto}).
 We use the NSGA II~\cite{deb2002fast}, an evolutionary algorithm that samples our forward BNN to discover the Pareto front iteratively.
 The uncertainty score in this experiment is the weighted sum of aleatoric and epistemic uncertainty.
 We set the values of the weights on the tuned values on the inversion task.
 The population size and the number of generations in this experiment are 1000 and 100, respectively.
Figure \ref{fig:pareto} suggests that the losses of uncertainty aware inversion are conflicting such that for example reducing the MSE loss will lead to the increase of the uncertainty score.

\section{Ablation of the number of networks in the ensemble}
\numberofnetworks
In this experiment we investigate the importance of the accuracy of the calculated uncertainties on the final NFP error.
We have trained deep ensemble networks with a varying number of networks in the ensemble.
The networks are trained on the spectral printer experiment for both noisy and sparse datasets.
As evident in Figure \ref{fig:numberofnetworks}, by increasing the number of networks in the ensemble the NFP error improves significantly. 
This trend indicates the importance of accurate prediction of the uncertainties.

\section{Implementation}
In practice, training the ensemble networks directly with negative log likelihood loss (Equation \ref{eq:NLL_loss}) is challenging \cite{seitzer2022pitfalls}.
Instead, following \cite{nix1994estimating}, we take a 3-step procedure for implementing deep ensemble predictive uncertainty.
We start with training an ensemble of conventional networks with diverse activation functions and MSE as its loss.
These networks are in fact the initialization of $\mathbb{F}_{\mathbf{\mu}}(\cdot)$.
The next step is training $\mathbb{F}_{\mathbf{\sigma}}(\cdot)$ and fine tuning $\mathbb{F}_{\mathbf{\mu}}(\cdot)$ jointly with the negative log likelihood loss (Equation \ref{eq:NLL_loss}).
Finally, we replace $\hat{f}(\cdot)$ with $\mathbb{F}_{\mathbf{\mu}}(\cdot)$  and incorporate $\mathbb{F}_{\mathbf{\sigma}}(\cdot)$ in the \texttt{Autoinverse} loss.

\end{document}